\documentclass{article}

 \usepackage[main, final]{neurips_2026}

\usepackage[utf8]{inputenc} 
\usepackage[T1]{fontenc}    
\usepackage{hyperref}       
\usepackage{url}            
\usepackage{booktabs}       
\usepackage{amsfonts}       
\usepackage{nicefrac}       
\usepackage{microtype}      
\usepackage{xcolor}         
\usepackage{graphicx}
\usepackage{booktabs}
\usepackage{wrapfig}
\usepackage{amsmath}
\usepackage{enumitem}
\usepackage{xspace}
\usepackage{caption}
\newcommand{\ie}{\textit{i.e.}\xspace}
\newcommand{\eg}{\textit{e.g.}\xspace}
\title{Vision to Geometry: 3D Spatial Memory for Sequential Embodied MLLM Reasoning and Exploration}

\author{%
  \makebox[0.4\textwidth][c]{Zhongyi Cai\thanks{Equal contribution.}} \\
  Michigan State University \\
  \texttt{caizhon2@msu.edu} \\
  \And
  \makebox[0.4\textwidth][c]{Yi Du\footnotemark[1]} \\
  University at Buffalo \\
  \texttt{yid@sairlab.org} \\
  \AND
  \makebox[0.4\textwidth][c]{Chen Wang} \\
  University at Buffalo \\
  \texttt{chenw@sairlab.org} \\
  \And
  \makebox[0.4\textwidth][c]{Yu Kong} \\
  Michigan State University \\
  \texttt{yukong@msu.edu} \\
}

\begin{document}

\maketitle

\begin{abstract}
Embodied agents are expected to assist humans by actively exploring unknown environments and reasoning about spatial contexts.
When deployed in real life, agents often face sequential tasks where each new task follows the completion of the previous one and may include infeasible objectives, such as searching for non-existent objects.
However, most existing research focuses on isolated goals, overlooking the core challenge of sequential tasks: the ability to reuse spatial knowledge accumulated from previous explorations to guide subsequent reasoning and exploration.
In this work, we investigate this underexplored yet practically significant embodied AI challenge.
Specifically, we propose 3DSPMR, a 3D SPatial Memory Reasoning framework that utilizes Field-of-View (FoV) coverage as an explicit geometric prior. 
By integrating FoV-based constraints, 3DSPMR significantly enhances an agent’s memory, reasoning, and exploration capabilities across sequential tasks.
To facilitate research in this area, we further introduce SEER-Bench, a novel Sequential Embodied Exploration and Reasoning Benchmark that spans two foundational tasks: Embodied Question Answering (EQA) and Embodied Multi-modal Navigation (EMN).
SEER-Bench uniquely incorporates both feasible and infeasible tasks to provide a rigorous and comprehensive evaluation of agent performance.
Extensive experiments verify that 3DSPMR achieves substantial performance gains on both sequential EQA and EMN tasks\footnote[1]{Our code is publicly available at 
\textcolor{magenta}{\url{https://github.com/zhyczy/3DSPMR}}.}.
\end{abstract}

\begin{figure*}[t]
\centering
  \includegraphics
  [width=\linewidth]
  {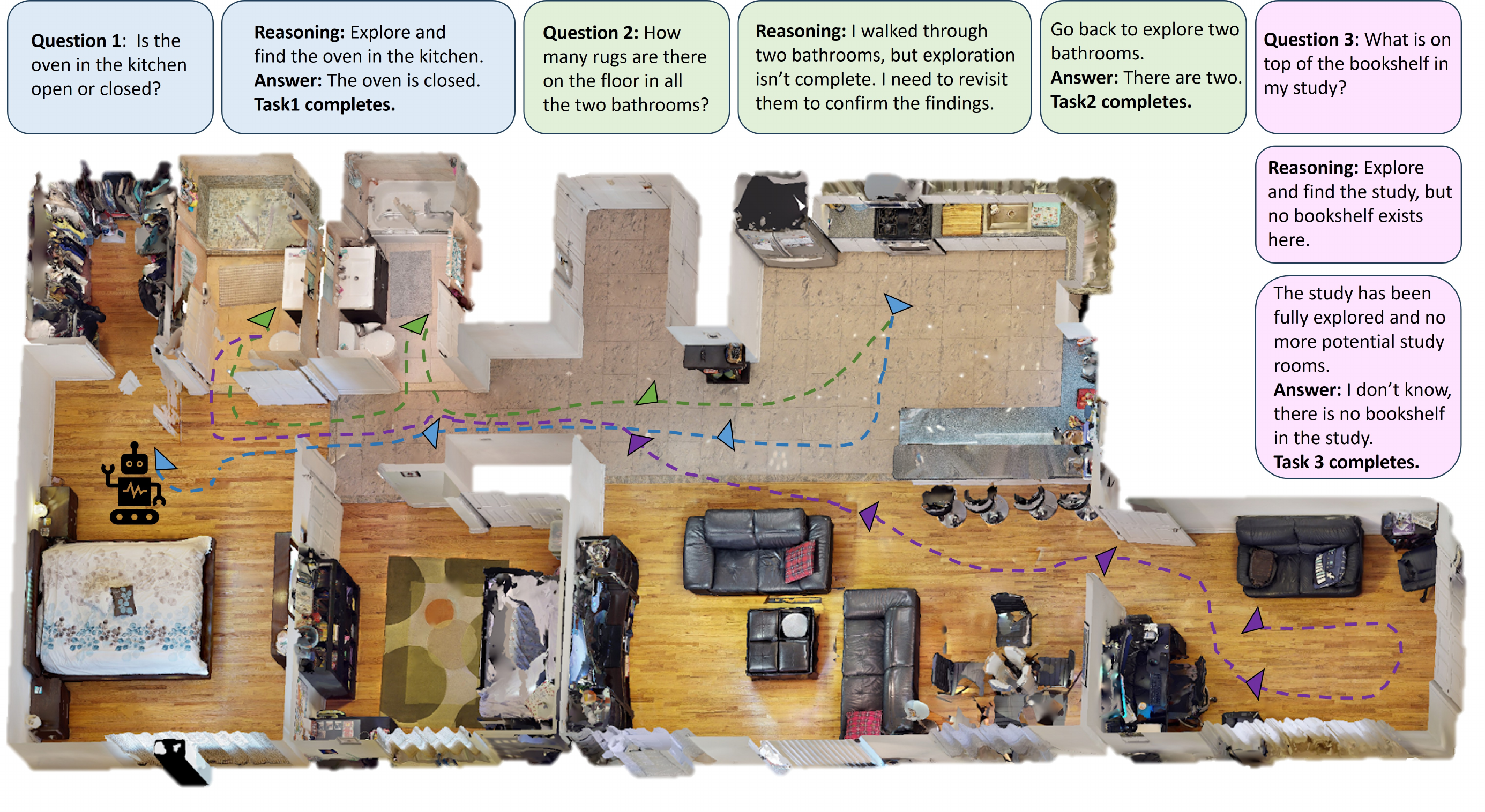}
  \caption{{\bfseries Illustration of 3DSPMR completing sequential EQA tasks.} After navigating to the kitchen and answering the first question, the agent receives the second query.
Using its spatial memory, it recognizes two previously passed but insufficiently explored bathrooms, revisits them, and answers the question after sufficient exploration.
For the third question, the agent locates the study but cannot find a bookshelf; once the study is fully explored and no alternative candidate rooms remain, 3DSPMR correctly identifies the task as infeasible.}
\vspace{-10pt}
\label{fig:task_illustration}
\end{figure*}

\section{Introduction}
\label{sec:intro}
In real-world scenarios, embodied agents frequently encounter sequential tasks, where tasks follow one another and may even be infeasible, such as searching for non-existent objects.
For feasible tasks, agents should operate efficiently; for infeasible ones, agents must detect the impossibility and communicate it clearly.
However, mainstream research in Embodied Question Answering (EQA)~\cite{li2025industryeqa, gordon2018iqa, ren2024explore, dang2025ecbench, linghu2024multi, wijmans2019embodied} and Embodied Multi-modal Navigation (EMN)~\cite{sgnav, batra2020objectnav, zhang20233d, yokoyama2024hm3d, long2024instructnav, zhou2024navgpt} remains largely confined to a single-task setting~\cite{majumdar2024openeqa, zhao2025cityeqa, jiang2025explorationeqa, deitke2020robothor}, where the agent's memory is reset for each isolated goal.
Only a few recent EMN studies~\cite{ khanna2024goat,song2025lhvln} have begun to consider the sequential setting, yet they still overlook infeasible tasks. 
To bridge this gap, we tackle the underexplored but critical problem of sequential embodied tasks, a crucial step toward developing more capable embodied agents.
The core challenge in this problem lies in enabling agents to effectively reuse spatial information accumulated from prior tasks for both feasible and infeasible objectives.

Multi-modal Large Language Models (MLLMs)~\cite{hurst2024gpt4o, comanici2025gemini} are powerful general-purpose models that have been applied to complex embodied AI problems, yet they still struggle with spatial understanding and reasoning~\cite{yang2025thinking, chen2025spatial}.
Although recent studies~\cite{3dmem, saxena2024grapheqa, sgnav} have employed memory-based reasoning frameworks to enhance embodied agents, these approaches still face three major challenges when applied to sequential embodied tasks.
First, their memory designs either omit essential cues or store excessive redundant details from previously explored areas during earlier tasks, causing agents to struggle with retrieving relevant information for the new task.
Second, MLLMs may overconfidently produce premature results or even hallucinations based on partial observations, due to incomplete exploration in earlier tasks.
Finally, these approaches fail to identify infeasible tasks, leading agents to repeatedly or endlessly explore already well-explored areas.

\begin{wrapfigure}{r}{0.35\textwidth} 
\vspace{-17pt}
  \centering
  \includegraphics[width=0.35\textwidth]{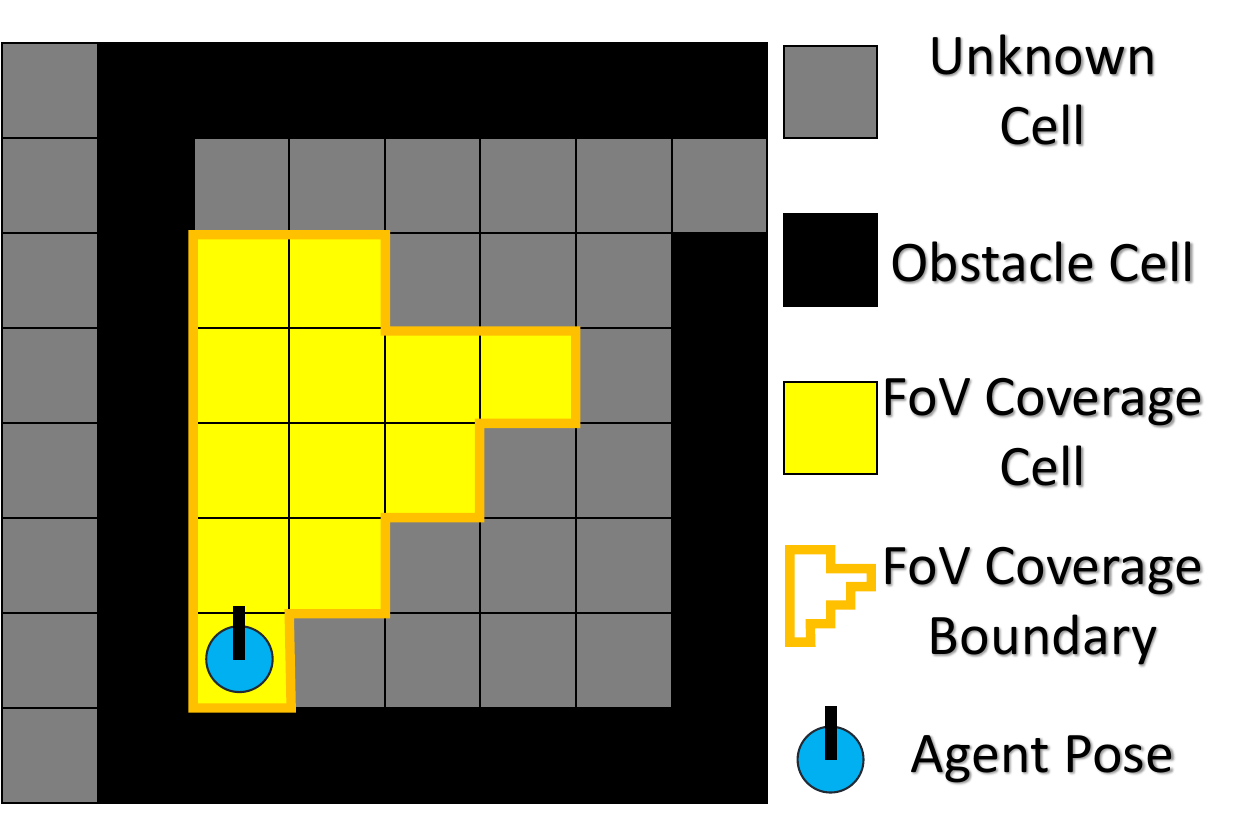}
  \caption{
  Top-down Projection of Agent FoV Coverage.
  }
  \label{fig:fov_figure}
\end{wrapfigure}
To address these challenges, we propose 3D SPatial Memory Reasoning (3DSPMR), which introduces explicit geometric cues derived from the Field-of-View (FoV) coverage to simultaneously enhance memory, reasoning, and exploration.
As illustrated in Fig.~\ref{fig:fov_figure}, FoV coverage represents the geometric region captured within the agent's viewing range. 
In this work, we perform ray casting from the agent's pose to identify visible cells in the top-down view. 
Those intersected by rays are marked as ``covered'' (yellow cells).
By modeling FoV coverage, 3DSPMR provides a structured geometric representation of exploration progress.

Specifically, our 3DSPMR framework enhances embodied agents across three critical dimensions. 
First, for memory, we propose a new FoV-based gating mechanism to decide when to save specific observations.
As a result, the memory keeps representative data and filters spatial redundancy. 
Second, for model reasoning, we propose to leverage FoV coverage to verify whether all task-relevant spatial information has been sufficiently collected. 
Through grounded geometric examination, this post-inference mechanism rectifies the MLLM’s predictions to mitigate common failure modes, such as premature task termination and the inability to recognize infeasible requests.
Finally, for exploration, we propose utilizing FoV coverage to incentivize exploration of underexplored regions.
Once triggered, the proposed Geometric and Semantic (Geo-Sem) module fuses geometric coverage with task-relevant semantics to identify optimal frontier targets, thereby improving navigation goal selection.

To systematically evaluate agents in these realistic scenarios and address the lack of suitable datasets, we construct the Sequential Embodied Exploration and Reasoning Benchmark (SEER-Bench). 
Unlike prior datasets, SEER-Bench covers both EQA and EMN in a sequential setting, rigorously evaluating an agent's ability to reuse spatial memory across a mixture of feasible and infeasible objectives.
Extensive experiments demonstrate that 3DSPMR substantially enhances both reasoning accuracy and exploration efficiency in previously unseen 3D environments.

Our contributions can be summarized as follows:
\begin{itemize}[leftmargin=*, nosep]
    \item 
    We propose a compact and representative spatial memory system that retains essential geometric cues and filters redundancy, effectively bridging the 3D scene understanding gap for MLLMs in sequential tasks.
    \item
    We develop the 3DSPMR framework, which leverages explicit geometric priors to guide both reasoning and exploration. 
    This enables agents to effectively execute sequential tasks and reliably identify infeasible objectives.
    \item 
    We propose SEER-Bench, a Sequential Embodied Exploration and Reasoning Benchmark that covers EQA and EMN, including both feasible and infeasible tasks to systematically evaluate agents' performance.
\end{itemize}
\section{Related Work}
\label{sec:related}

\noindent\textbf{MLLMs for Embodied and Spatial Reasoning.}
Recent Multimodal Large Language Models (MLLMs) such as GPT-4o~\cite{hurst2024gpt4o}, Qwen2VL~\cite{wang2024qwen2vl}, and LLaVA-OneVision~\cite{li2024llavaonevision} have shown remarkable progress in general visual–language reasoning. 
These models have inspired growing efforts toward embodied and spatial reasoning, where agents interpret 3D environments and perform goal-directed behaviors~\cite{mu2023embodiedgpt, goetting2025vlmnav, liu2024llavaplus}.
However, most existing MLLMs rely on 2D image–text supervision and lack explicit 3D spatial awareness~\cite{yang2025thinking, chen2025spatial}.
This limitation constrains their ability to reason about geometry, occlusion, and room layouts, which are crucial for embodied perception and navigation.
Moreover, their frame-wise visual processing hinders persistent memory formation, leading to fragmented and inconsistent spatial understanding in long-horizon tasks.
As a solution, our proposed 3DSPMR approach leverages a unified spatial memory to distill complex 3D representations into text-form scene graphs, images, and a geometric map, providing a more accessible input format for MLLMs.

\noindent\textbf{3D Representations and  Memory in Embodied Agents.}
Structured spatial representations are central to effective reasoning and interaction within 3D environments. 
Scene-graph-based methods~\cite{gu2024conceptgraphs, werby2024hovsg, hughes2022hydra, hughes2024foundations, rana2023sayplan} encode object relationships and spatial hierarchies, offering interpretable relational abstractions.
However, such symbolic graphs are typically sparse and lack the visual richness needed for perceptual grounding.
Complementary approaches that store egocentric visual memories~\cite{3dmem, yang2024emma, chang2023goat} allow retrieval of past observations but often lack geometric or relational awareness.
Recent methods~\cite{liu2024ver, 3dmem, du2025vl} have explored constructing 3D semantic maps and global spatial memories, which enhance navigation and question answering yet treat relational, visual, and geometric cues independently.
In contrast, 3DSPMR integrates these complementary modalities into a unified 3D spatial memory, combining relational structure, visual appearance, and geometric coverage for coherent spatial reasoning.

\noindent\textbf{Sequential Embodied Tasks.}
Traditional embodied research such as Embodied Question Answering (EQA)~\cite{das2018embodied, yu2019multi} and Scene-Graph Navigation (SG-Nav)~\cite{sgnav} evaluate isolated single-task scenarios where each episode is independent.
Recent works have explored sequential and multi-task settings~\cite{song2025lhvln, yin2025unigoal, zheng2024navillm}, but they typically assume all goals or questions are feasible.
In realistic environments, some queries are inherently \textit{infeasible}, such as asking for the color of a non-existent object or navigating to an unreachable area.
Addressing such cases requires feasibility awareness in addition to spatial reasoning—an ability largely overlooked in prior benchmarks.
Recent extensions of EQA, such as OpenEQA~\cite{majumdar2024openeqa}, GraphEQA~\cite{saxena2024grapheqa}, and exploration-aware EQA~\cite{jiang2025explorationeqa}, have begun to tackle more diverse and dynamic settings but still do not explicitly incorporate infeasible task reasoning.
Our SEER-Bench extends this paradigm by including both feasible and infeasible tasks in sequential EQA and EMN episodes.
Therefore, SEER-Bench extends the embodied task formulation, offering a more realistic and comprehensive evaluation for real-world embodied agents.

\section{Methodology}
\label{sec:methodology}
\begin{figure*}[t]
\centering
  \includegraphics
  [width=\linewidth]
  {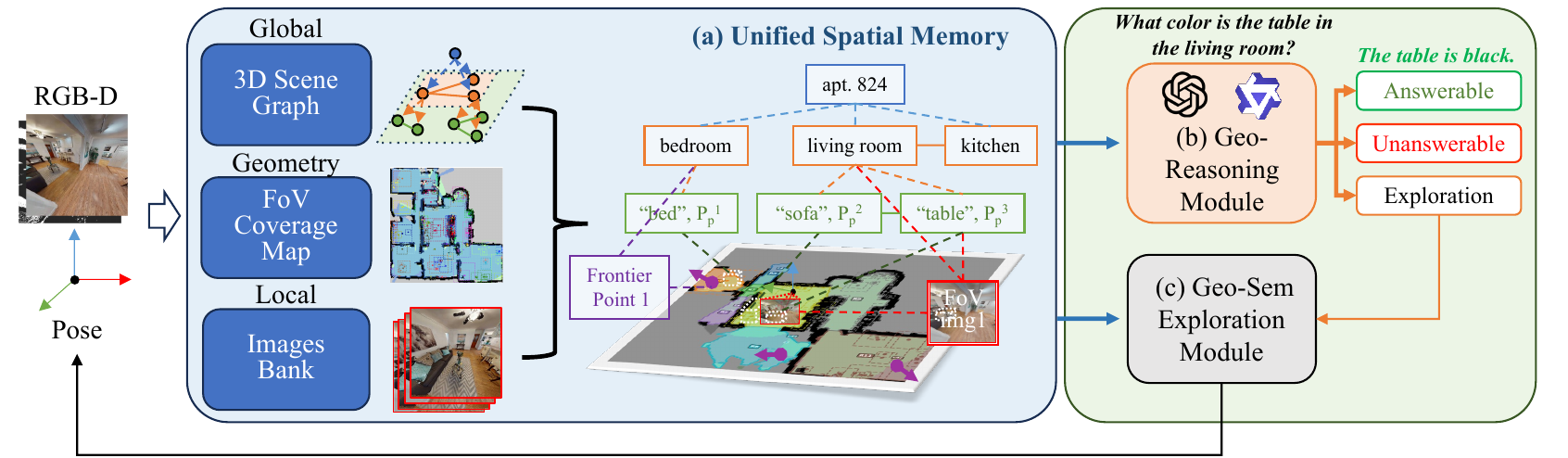}
  \caption{\textbf{Overview of 3DSPMR.} 
  (a) Unified Spatial Memory: Raw observations are selectively stored to maintain an informative structured spatial representation. 
  (b) Geo-Reasoning: A grounded inference module that ensures robust decision-making.
(c) Geo-Sem Exploration: A strategic module that identifies optimal frontier targets.
}
\vspace{-10pt}
\label{fig:pipeline}
\end{figure*}
We introduce 3DSPMR, a framework designed to enhance memory retention and feasibility awareness in sequential embodied tasks.
As exhibited in Fig.~\ref{fig:pipeline}, 3DSPMR consists of three synergistic components: a unified spatial memory, a Geo-Reasoning module, and a Geo-Sem Exploration module. 

\subsection{Unified Spatial Memory}
\label{subsec:memory}
In sequential embodied tasks, agents must be able to consolidate and reuse historical information accumulated across multiple tasks, which necessitates a compact yet representative memory.
However, existing memory architectures often exhibit two extremes: they either focus exclusively on the immediate task~\cite{ren2024explore, saxena2024grapheqa}, or suffer from excessive spatial redundancy~\cite{3dmem, wang2025think}.
Consequently, when transitioning to new tasks, agents may either lack essential context or become encumbered by irrelevant information.

To address this issue, 3DSPMR maintains a unified spatial memory that continuously integrates representative information from all observations. 
These compact, persistent representations enable the agent to either directly facilitate reasoning, such as answering EQA queries from memory, or guide exploration by leveraging previously accumulated environmental cues.

\begin{wrapfigure}{r}{0.4\textwidth} 
\vspace{-15pt}
  \centering
  \includegraphics[width=0.4\textwidth]{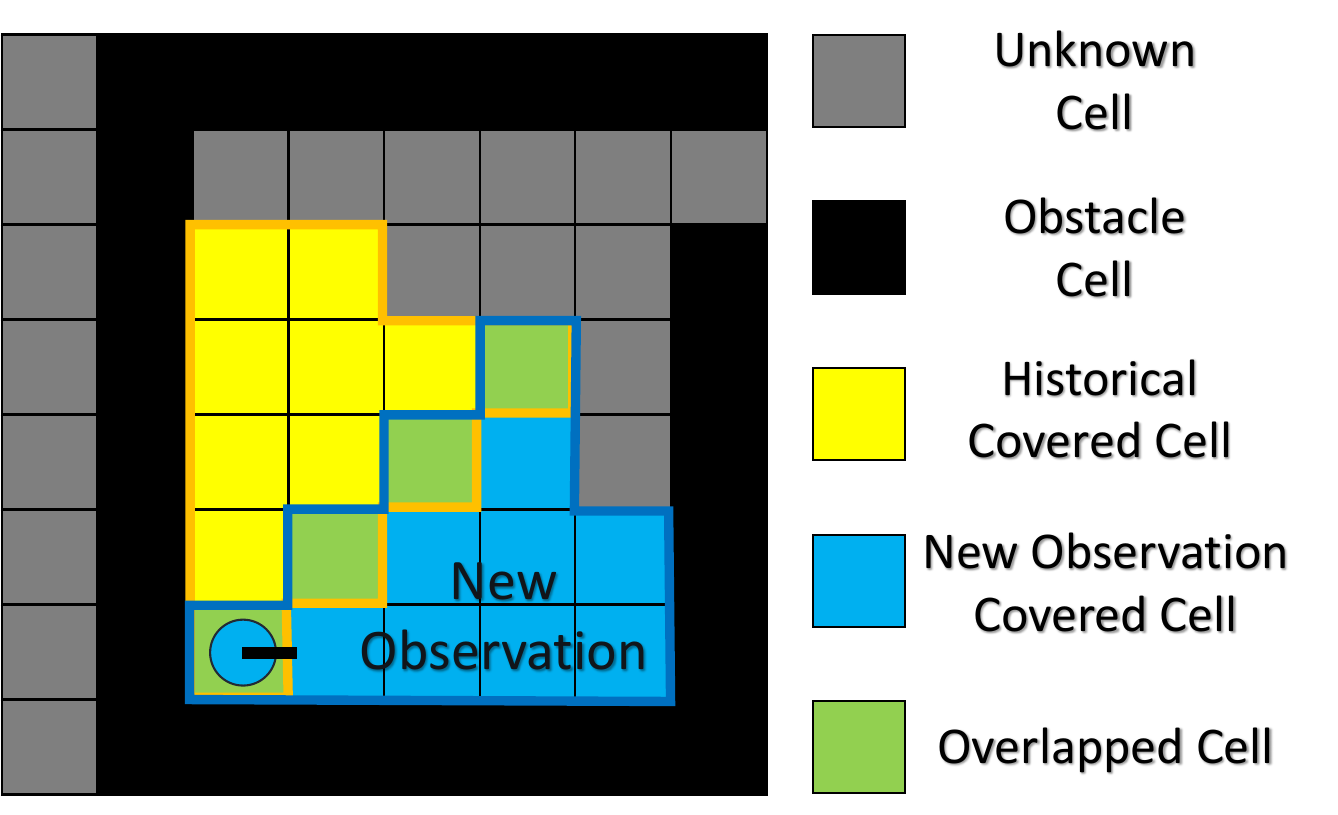}
  \caption{
  \textbf{
  Top-down Projection of FoV-based Gating.
  }
  After a $90^{\circ}$ rotation, the agent obtains a new observation. Using the historical FoV coverage (yellow cells) and the coverage from the new observation (blue cells), we calculate the overlap (green cells) to decide whether to save this observation.
  } 
  \label{fig:fov_coverage_figure}
\end{wrapfigure}

As shown in Fig.~\ref{fig:pipeline}(a), the unified spatial memory comprises three complementary modalities: global relational cues that provide topological context via a 3D scene graph, local visual cues that preserve high-fidelity visual attributes through an ego-centric image bank, and geometric cues derived from a Field-of-View (FoV) coverage map.
Crucially, these geometric cues serve as a grounded anchor for the model to assess task feasibility and completion while seamlessly linking local and global representations.
Furthermore, we employ a FoV-based gating mechanism to retain only non-redundant observations, significantly reducing the storage overhead typically incurred by resource-intensive visual data.

\noindent{\bf Global Relational and Local Visual Cues.}
Globally, we represent the environment as a hierarchical 3D scene graph $G$ comprising \textit{Apartment}, \textit{Room}, and \textit{Object} levels. 
We first build upon ConceptGraphs~\cite{gu2024conceptgraphs} to maintain object nodes, and then insert room nodes through morphological processing of the top-down projected occupancy map.
Although this graph provides a highly interpretable and structured topological map, it inevitably omits fine-grained visual textures and high-fidelity appearance details. 
To compensate for this, we maintain a keyframe image bank $I$ as a repository for high-resolution egocentric observations. 
The geometric cues described below determine the timing and selection of image storage to ensure that $I$ remains both compact and representative.

\noindent{\bf Geometric Cues.}
Unlike previous memory designs, we introduce FoV coverage (shown in Fig.~\ref{fig:fov_figure}) as a fundamental geometric cue. 
Based on this cue, we propose a geometric gating mechanism to maintain a compact image memory. 
We contend that visual observations should only be stored when they yield substantial geometric expansion, thereby enriching the memory with sufficiently novel spatial information.
We quantify geometric novelty as the proportion of the current view that encompasses previously unobserved spatial regions. 
Specifically, given the current FoV coverage $C_{new}$ (the blue boundary region in Fig.~\ref{fig:fov_coverage_figure}) and the accumulated historical coverage $C_{history}$ (the yellow boundary region in Fig.~\ref{fig:fov_coverage_figure}), the novelty score $\tau_{novel}$ is computed as:
\begin{equation}
\label{eq:novelty}
\tau_{novel} = 1 - \frac{|C_{new} \cap C_{history}|}{|C_{new}|}.
\end{equation}
At step $t$, if $\tau_{novel}$ exceeds a predefined threshold $\tau_{0}$, the system archives this image in the image bank $I_t$ and updates the coverage map via $C_t = C_{history} \cup C_{new}$. 
This union incorporates the newly observed cells (blue cells in Fig.~\ref{fig:fov_coverage_figure}).
This strategy naturally leads to memory saturation once the environment is fully explored, effectively filtering redundant views while preserving a representative visual history for sequential tasks.

Beyond filtering, FoV coverage serves as a geometric bridge to establish bidirectional links between keyframe images and 3D graph nodes.
Specifically, each saved image in $I$ is associated with 
\begin{wrapfigure}{r}{0.4\textwidth} 
\vspace{-9pt}
  \centering
 \includegraphics[width=0.4\textwidth]{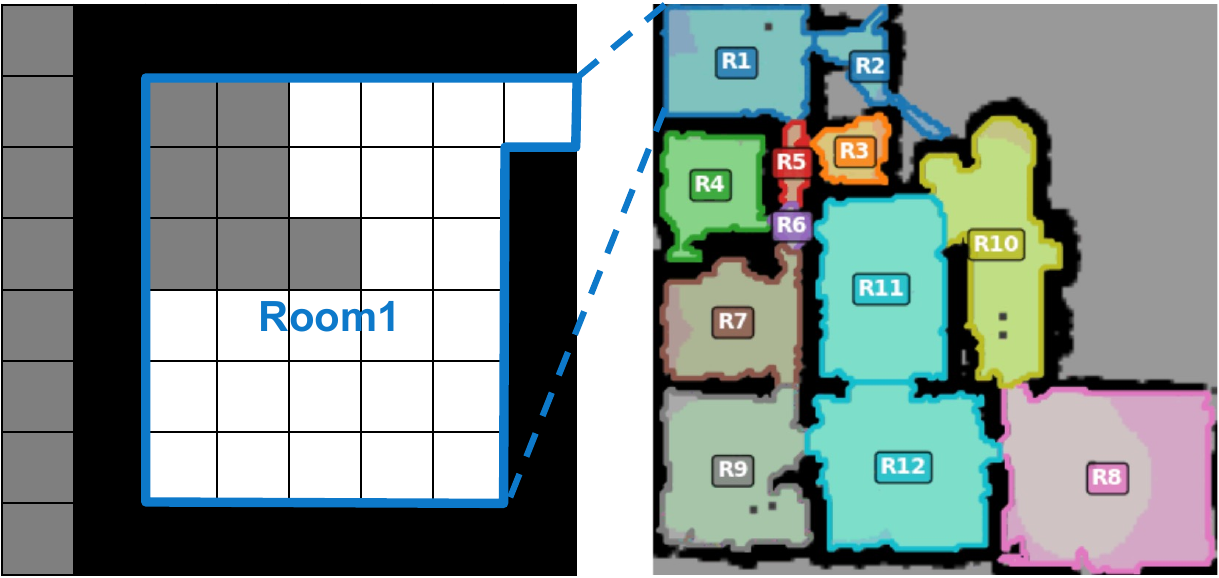}
  \caption{\textbf{
  Top-down Projection of Room Coverage.
  } The blue boundary defines the area of Room 1, while white cells mark covered regions $\mathrm{C_{history}}$. The room coverage ratio is calculated as the fraction of the total room area $\mathrm{S_{room}}$ (cells within the boundary) that has been covered by the agent.}
  \label{fig:room_covereage}
\end{wrapfigure}
the object nodes within its visibility frustum. 
Conversely, each object node stores references to all images in which it is captured. 
This cross-modal indexing enables seamless transitions between global and local cues, synthesizing them into a unified memory architecture.

Finally, the FoV coverage provides a quantitative metric for exploration progress. 
By intersecting the historical FoV coverage $C_{history}$ with the whole area of a specific room $S_{room}$ (\eg, the area of Room 1 enclosed by the blue boundary in Fig.~\ref{fig:room_covereage}), we derive the coverage rate $\eta$ to represent the explored proportion of this room:
\begin{equation}
\label{eq:coverage}
\eta = \frac{|C_{history} \cap S_{room}|}{|S_{room}|}.
\end{equation}
This metric provides grounded geometric evidence for the agent to assess each room's exploration status and serves as a termination criterion for task completion.

\begin{figure}[t]
\centering
  \includegraphics
  [width=0.9\linewidth]
  {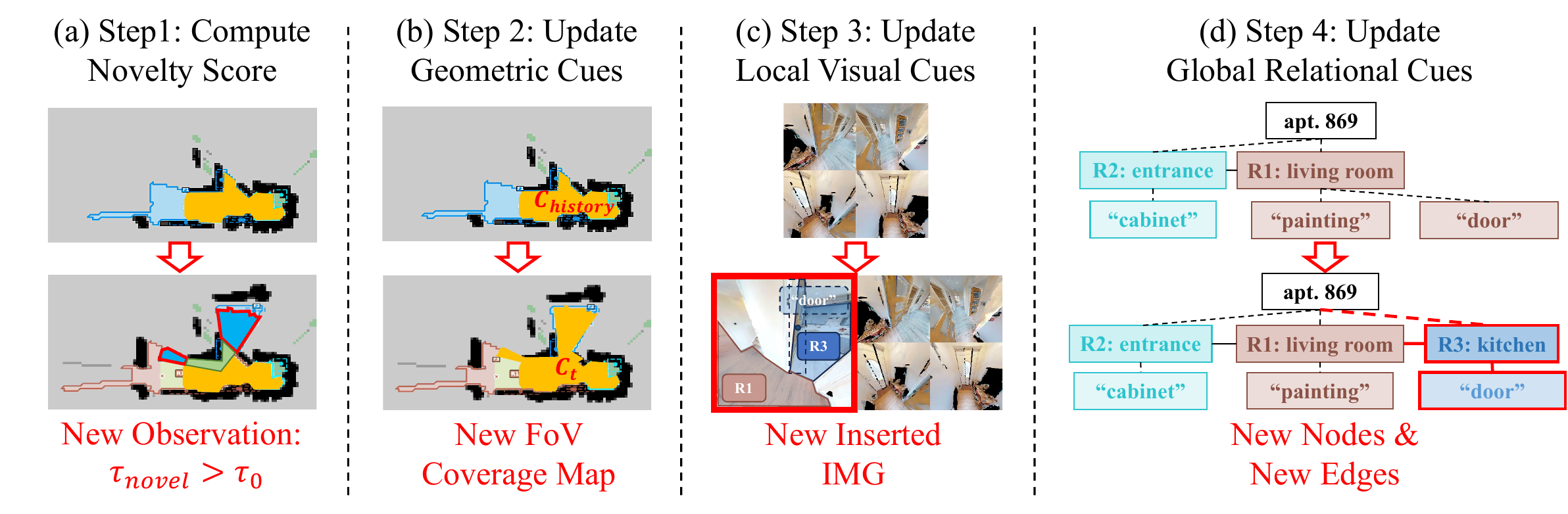}
  \caption{\textbf{Unified Spatial Memory Update.} 
  The update process comprises four steps. 
  (a) Step 1: compute the novelty score $\tau_{\!novel}$ of the new observation. 
  If $\tau_{\!novel}>\tau_0$, update memory in later steps. 
  (b) Step 2: 
  update geometric cues by merging the historical coverage map with the new image's FoV coverage.
  (c) Step 3: update local visual cues by saving this image into the image bank.
  (d) Step 4: update global relational cues by updating the graph and establishing bidirectional links between the graph and the image.
  }
  \vspace{-14pt}
\label{fig:memory}
\end{figure}

\noindent{\bf Unified Spatial Memory.}
Together, these cues form the unified spatial memory at step $t$, defined as $\mathcal{M}_t=f_{\mathrm{3DSPMR}}(G_t,\,
                     I_t,\,
                     C_t)$,
where the global 3D scene graph $G_t$ encodes symbolic structure, the local image bank $I_t$ provides perceptual grounding, and the FoV coverage map $C_t$ maintains geometric context.
At each step $t$, the agent acquires RGB-D observations from its current pose to perceive the local environment. 
These data are subsequently integrated into the unified spatial memory, as illustrated in Fig.~\ref{fig:memory}.
The system first computes the geometric novelty score $\tau_{\!novel}$ (Eq.~\ref{eq:novelty}) to decide whether to update $C_t$ and to archive the observation frame in $I_t$.
If the frame is stored, the scene graph is updated accordingly, and bidirectional links are established between the new image and the relevant object nodes. 
By synchronizing these heterogeneous cues, 3DSPMR bridges the gap in 3D scene understanding for MLLMs, providing a robust foundation for downstream reasoning.

\subsection{Geo-Reasoning}
\label{subsec:reasoning}
Building on the unified spatial memory, we introduce a Geo-Reasoning module designed to utilize multi-modal cues for task execution and infeasibility detection. 
Although MLLMs exhibit remarkable perceptual and reasoning capabilities when processing text and image inputs, they are prone to reaching premature conclusions in partially explored environments and falling into repetitive loops when facing infeasible tasks. 
Unlike prior approaches~\cite{ren2024explore, jiang2025explorationeqa, saxena2024grapheqa} that rely on model-based confidence heuristics, which often suffer from calibration issues, our Geo-Reasoning module leverages model-agnostic geometric information (\ie, FoV coverage) to provide a grounded, external verification.

Specifically, we introduce a Geometric Examination Mechanism to perform post-hoc validation of MLLM predictions.
This mechanism leverages the cumulative coverage map $C$ stored in the spatial memory to assess exploration completeness.
If the coverage rate $\eta$ (Eq.~\ref{eq:coverage}) of all task-relevant rooms exceeds a predefined threshold, the inference is deemed grounded and accepted. 
Conversely, if underexplored relevant regions persist, the mechanism rejects the inference and triggers further exploration.
For accepted cases where a definitive answer is provided, the agent commits to it as the final prediction.
Otherwise, the mechanism examines the coverage of irrelevant rooms to identify any unobserved yet potentially relevant areas. 
If no unexplored areas remain, the task is formally classified as infeasible.
If unexplored regions persist, the system initiates Geo-Sem exploration to gather additional information.
By bridging geometric occupancy with semantic reasoning, this mechanism effectively prevents premature task termination, detects infeasible tasks, and mitigates redundant revisiting of explored space. 
Details and visual illustrations of the reasoning process are provided in the Appendix.
\subsection{Geo-Sem Exploration}
\label{subsec:exploration}
In scenarios necessitating exploration, the agent must identify the most informative unexplored region to visit next.
Prior works~\cite{majumdar2024openeqa, yin2025unigoal} primarily leverage visual semantics from frontier images for navigation; however, such a paradigm ignores critical geometric considerations, which may result in inefficient trajectories, especially when semantic signals alone fail to distinguish between multiple candidates.

\begin{wrapfigure}{r}{0.35\textwidth} 
\vspace{-23pt}
  \centering
  \includegraphics[width=0.35\textwidth]{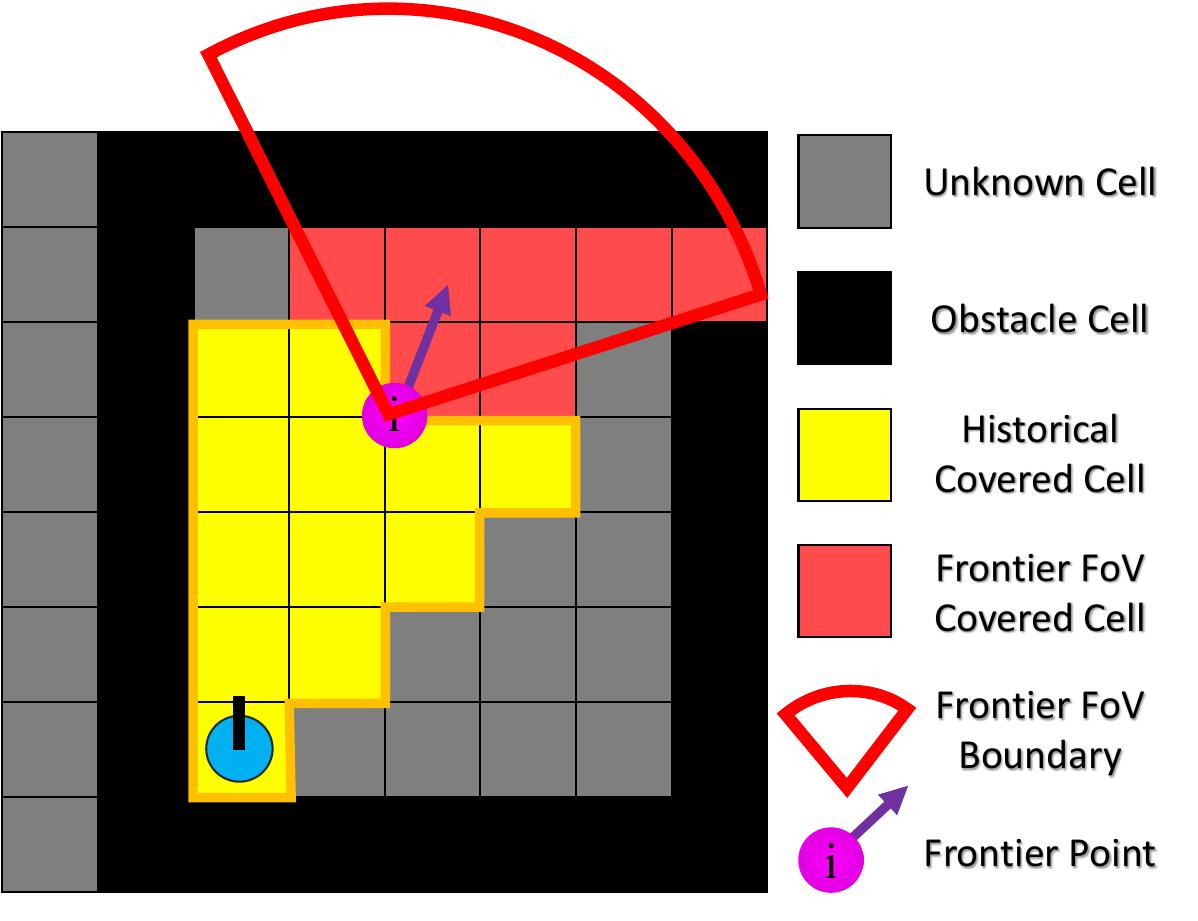}
  \caption{
  \textbf{
  Top-down Projection of Frontier Exploration FoV.
  }
  The red lines delineate the FoV projected ahead of a frontier node. 
  Red cells indicate unknown cells within the FoV, whose area is denoted as $U^{\mathrm{FoV}}_t$ in Eq.~\ref{eq:frontier_scores_fov}.
  }
  \label{fig:fov_coverage_extended}
\end{wrapfigure}
The proposed Geo-Sem Exploration module addresses these limitations by integrating geometric cues with visual semantics, enabling the agent to holistically determine the next target region.
At each exploration step, the occupancy map identifies a set of frontiers $\mathcal{F}_t$, defined as the boundary regions between explored (yellow cells) and unexplored unknown space (gray cells). 
For each frontier node $i \in \mathcal{F}_t$ (the purple point shown in Fig.~\ref{fig:fov_coverage_extended}), Geo-Sem leverages an MLLM $r$ to assign a semantic relevance score $v^{\mathrm{sem}}_t(i)$  conditioned on the task query $Q$ and the frontier's visual context in node $i$, defined as: $v^{\mathrm{sem}}_t(i) = r(i, Q)$.
This score quantifies the potential of frontier node $i$ to yield information crucial for fulfilling the current task.

Additionally, Geo-Sem utilizes two geometric heuristics to guide exploration. 
Most notably, frontier node $i$'s FoV is used to evaluate the potential unobserved area beyond it.
Based on this, Geo-Sem computes the exploration potential score $v^{\mathrm{exp}}_t(i)$ as follows:
\begin{equation}
\label{eq:frontier_scores_fov}
v^{\mathrm{exp}}_t(i) = \frac{U^{\mathrm{FoV}}_t(i)}{U_{\max}},
\end{equation}
where $U^{\mathrm{FoV}}_t(i)$ denotes potential explorable area (red cells in Fig.~\ref{fig:fov_coverage_extended}), and $U_{\max}$ denotes the largest possible explorable area (the area within the whole red boundary).
By incentivizing the agent to prioritize regions with larger unexplored area, our strategy expedites the acquisition of a holistic environmental representation, effectively preventing suboptimal motion patterns such as repetitive oscillations or excessive detours. 
To favor proximal targets, Geo-Sem incorporates geometric proximity to the frontier as a secondary heuristic.
This proximity score, $v^{\mathrm{dist}}_t(i)$, is formulated as: 
$v^{\mathrm{dist}}_t(i) = 1/(1 + \|p_i - p_t\| / d_{\max}),$
where $p_t$ is the agent's current position, $p_i$ is the frontier pos, and $d_{\max}$ serves as a distance normalization constant.
This refinement enables the agent to select closer frontiers, particularly when the aforementioned heuristics yield comparable scores across multiple candidates.

Finally, these components are integrated into a unified composite score $V_t(i)$ as: $V_t(i) = w_{\mathrm{sem}}\,v^{\mathrm{sem}}_t(i) + w_{\mathrm{exp}}\,v^{\mathrm{exp}}_t(i) + w_{\mathrm{dist}}\,v^{\mathrm{dist}}_t(i),$
with weighting coefficients $(w_{\mathrm{sem}}, w_{\mathrm{exp}}, w_{\mathrm{dist}})$ normalized to sum to 1. 
The target frontier is selected by maximizing the composite score: $i^{\star} = \arg\max_{i \in \mathcal{F}_t} V_t(i).$

Once the frontier $i^{\star}$ is selected, the agent navigates toward its location $p_{i^{\star}}$ using a shortest-path planner within the explored region. 
Such a multi-faceted integration enables the agent to comprehensively determine the next-best-step, significantly enhancing exploration efficiency through the synergy of semantic intent and geometric viability.
In practice, we employ the pathfinder from Habitat-Sim~\cite{szot2021habitat} to compute such routes and to advance the agent by small steps at each simulation tick.

\section{SEER Benchmark}
\label{sec:seer-bench}
\noindent{\bf Overview.} Existing benchmarks~\cite{majumdar2024openeqa, ren2024explore, deitke2020robothor} are mainly designed for single-task cases, which typically provide only one or several tasks per scenario and lack infeasible objectives, rendering them insufficient for a systematic evaluation of sequential embodied reasoning. 
To bridge this gap, we introduce the Sequential Embodied Exploration and Reasoning Benchmark (SEER-Bench). 
We define an exploration episode as the comprehensive process of exploring an environment and attempting all assigned tasks.
By augmenting each scenario with additional annotations, SEER-Bench supports multi-task evaluation within a single episode and assesses an agent's ability to reuse spatial knowledge and detect infeasible objectives.

SEER-Bench incorporates two core components: Embodied Question Answering (EQA) and Embodied Multi-modal Navigation (EMN), across 60 scenarios developed using HM3DSem~\cite{yadav2023habitat} within the Habitat simulator~\cite{szot2021habitat}.
Each scenario comprises 5 to 10 sequential tasks, 40$\%$ of which are infeasible, providing a realistic foundation for long-horizon sequential embodied reasoning. 
Comprehensive details regarding annotations and data statistics are provided in the Appendix.

\noindent\textbf{Evaluation Metrics.} 
Let $\mathcal{E}$ denote the set of test episodes, where each episode $e \in \mathcal{E}$ comprises a sequence of $N_e$ tasks.
For EMN, the agent must navigate to a target goal for each individual task $i$.
We define the success of task $i$ using a binary indicator $S_i \in \{0, 1\}$. 
For each task $i$, let $l_i$ denote the shortest path distance from the agent's initial position at the onset of the task to the goal, and let $h_i$ represent the actual trajectory length traversed by the agent during that specific task.

Standard evaluation metrics, such as Success Rate ($SR = \frac{1}{N} \sum_{i=1}^{N} S_i$) and Success-weighted Path Length ($SPL = \frac{1}{N} \sum_{i=1}^{N} S_i \frac{l_i}{\max(h_i, l_i)}$)~\cite{anderson2018evaluation}, where $N = \sum_{e \in \mathcal{E}} N_e$ denotes the total number of tasks, effectively measure the correctness and efficiency of isolated single tasks. 
However, they fail to capture task interdependencies or the efficiency of spatial knowledge reuse across an episode.
While one could simply average the SR over tasks within an episode, such an approach often obscures critical failures within the episode.
To address these limitations, we introduce two new metrics that impose strict requirements on long-horizon performance: Sequential Success Rate (SSR) for sequential correctness and Sequential SPL (SSPL) for sequential efficiency.

SSR measures the reliability of an agent across the entire episode. Unlike the average SR, which may award partial credit for completing individual steps, SSR serves as a strict binary metric for the entire episode. 
We define the success of an episode $e$ using a binary indicator $SR_e \in \{0, 1\}$, which equals 1 if and only if every task within that episode is successfully completed ($S_i = 1$ for all $i \in \{1, \dots, N_e\}$). 
Thus, SSR reflects the proportion of total episodes $\mathcal{E}$ that achieve full task completion: $\text{SSR}_{\text{EMN}} = \frac{1}{|\mathcal{E}|} \sum_{e \in \mathcal{E}} \mathbb{I}[\text{SR}_e = 1]\times 100\%.$

SSPL evaluates how efficiently an agent utilizes its total step budget to complete an entire episode. 
Crucially, this metric serves as a proxy for knowledge reuse: in sequential tasks, maximizing efficiency requires the agent to recall previously explored areas to accelerate subsequent goals, rather than redundantly re-exploring the environment.
Let $H_e$ denote the total steps taken by the agent during episode $e$, and let $L_e$ represent the step budget allocated for that episode.
SSPL rewards agents that achieve complete sequential success ($SR_e = 1$) while conserving their step budget: $\text{SSPL}_{\text{EMN}} = \frac{1}{|\mathcal{E}|} \sum_{e \in \mathcal{E}} \left( \mathbb{I}[SR_e = 1] \times \frac{L_e}{\max(H_e, L_e)} \right) \times 100\%.$

For EQA, we also introduce SSR and SSPL to evaluate model performance over a full episode. 
Their formulas, along with the setting of the step budget $L_e$, are provided in the Appendix.

\section{Experiment}
\label{sec:exp}
\label{subsec:exp_set}
\noindent\textbf{Setup.}
Using SEER-Bench, we evaluated an agent's capacity for multi-step reasoning and active exploration across both the EQA and EMN tracks. 
For experimental setup, we constructed a testing set by sampling 5 tasks from the annotated tasks in each of the 60 scenarios. 
Specifically, for each scenario, we randomly selected three feasible and two infeasible tasks. 
These five tasks were then shuffled to create a sequential task chain, resulting in a total of 300 tasks in each track.
The EQA track maintains a wide diversity of question types which include functional, counting, and spatial relationships, and incorporates both answerable and unanswerable queries to test the agent's discernment. 
The EMN track features diverse goal modalities, requiring the agent to navigate based on Object, Language, or Image targets.

\noindent{\bf Baselines.}
We benchmarked our approach against several representative methods across the EQA and EMN tracks to validate the efficacy of 3DSPMR. 
Specifically, we compared our approach with four state-of-the-art memory-enhanced active exploration methods: Explore-Mem~\cite{ren2024explore}, 3D-Mem~\cite{3dmem}, GraphEQA~\cite{saxena2024grapheqa}, and PredEQA~\cite{yuan2026predict}. 
To ensure a fair comparison, all baselines utilize Qwen2.5VL-72B~\cite{bai2025qwen2} as their default backbone across both tracks.
To isolate the influence of geometric information within our proposed 3DSPMR, we conducted an ablation study (w/o Geo) by removing the FoV-based geometric examination mechanism during reasoning and the geometric heuristics during exploration. 
For the EQA track, we additionally incorporated Blind LLMs, specifically Qwen3-14B~\cite{yang2025qwen3} and GPT-5~\cite{openai2025gpt5}, to serve as a lower bound by attempting to answer questions without active navigation.
In the EMN track, we further evaluated two distinct architectures: Modular GOAT~\cite{chang2023goat}, a zero-shot modular baseline lacking persistent memory, and SenseAct-NN~\cite{khanna2024goat}, a monolithic RL baseline. 
For a rigorous and robust assessment, we benchmarked the QA performance across multiple LLM evaluators, specifically GPT-5~\cite{openai2025gpt5}, Gemini-3.6-Flash~\cite{deepmind2026gemini36flash}, and LLaMA-3.3-70B~\cite{meta2024llama33}.

\begin{table}[htbp]
\centering
\caption{SEER-Bench EQA Results.}
\label{tab:eqa_results}
\resizebox{\linewidth}{!}{
\begin{tabular}{l cccc cccc cccc}
\toprule
& \multicolumn{4}{c}{GPT-5} & \multicolumn{4}{c}{Gemini-3.6-Flash} & \multicolumn{4}{c}{LLaMA-3.3-70B} \\
\cmidrule(lr){2-5}\cmidrule(lr){6-9}\cmidrule(lr){10-13}
Method & SSR $\uparrow$ & SSPL $\uparrow$ & SR $\uparrow$ & SPL $\uparrow$ & SSR $\uparrow$ & SSPL $\uparrow$ & SR $\uparrow$ & SPL $\uparrow$ & SSR $\uparrow$ & SSPL $\uparrow$ & SR $\uparrow$ & SPL $\uparrow$ \\
\midrule
\multicolumn{13}{l}{\textit{Blind LLMs}} \\
\midrule
Qwen3 & 0.7 & $\backslash$ & 26.0 & $\backslash$ & 0.3 & $\backslash$ & 26.2 & $\backslash$ & 0.7 & $\backslash$ & 28.9 & $\backslash$ \\
GPT-5 & 1.8 & $\backslash$ & 26.6 & $\backslash$ & 1.7 & $\backslash$ & 27.5 & $\backslash$ & 1.5 & $\backslash$ & 28.3 & $\backslash$ \\
\midrule
\multicolumn{13}{l}{\textit{Open-Sourced Qwen2.5VL Exploration}} \\
\midrule
Explore-Mem & 0.3 & 0.2 & 22.1 & 8.5 & 0.2 & 0.1 & 22.1 & 8.6 & 0.2 & 0.1 & 23.2 & 8.8 \\
GraphEQA & 1.5 & 0.4 & 19.8 & 9.1 & 1.2 & 0.3 & 20.0 & 9.4 & 2.2 & 0.7 & 32.4 & 19.1 \\
3D-Mem & 1.2 & 0.6 & 31.2 & 19.5 & 2.7 & 1.5 & 33.1 & 21.0 & 3.2 & 1.9 & 35.2 & 21.9 \\
PredEQA & 2.3 & 2.1 & 33.5 & 22.7 & 4.2 & 3.4 & 35.7 & 25.1 & 5.7 & 4.2 & 40.3 & 28.2 \\
3DSPMR w/o Geo & 5.8 & 4.9 & 39.1 & 27.7 & 6.3 & 5.4 & 40.8 & 29.2 & 8.8 & 7.6 & 46.6 & 33.1 \\
\textbf{3DSPMR (Ours)} & \textbf{14.5} & \textbf{12.9} & \textbf{48.8} & \textbf{41.3} & \textbf{12.8} & \textbf{11.4} & \textbf{49.1} & \textbf{41.6} & \textbf{16.2} & \textbf{14.4} & \textbf{51.9} & \textbf{44.8} \\
\bottomrule
\end{tabular}}
\end{table}

\textbf{Discussion and Analysis.}
We present a quantitative analysis of SEER-Bench results, summarized in Table~\ref{tab:eqa_results} (EQA) and Table~\ref{tab:emn_results} (EMN). 
3DSPMR consistently outperforms all baselines across every metric on both tracks. 
On the EQA track, the results remain consistent across different LLM judges, proving that our method's superiority is robust to the choice of evaluator.
Notably, our framework yields gains of approximately 10 to 20 percentage points on the rigorous SSR and SSPL metrics over 3D-Mem~\cite{3dmem} across both EQA and EMN tracks.

Standard baselines struggle with the sequential demands of our benchmark.
In particular, Explore-Mem~\cite{ren2024explore} confines memory to the current task, precluding the reuse of historical context for subsequent queries. 
Although 3D-Mem~\cite{3dmem}, GraphEQA~\cite{saxena2024grapheqa}, and PredEQA~\cite{yuan2026predict} maintain historical memory, they are still inferior to 3DSPMR in both task-level (SR, SPL) and episode-level (SSR, SSPL) metrics. 
This underperformance is due to their inability to both explicitly detect infeasible objectives and prevent premature task completion.

\begin{wraptable}{r}{0.5\textwidth}
    \vspace{-14pt}
    \centering
    \caption{SEER-Bench EMN Results.}
    \label{tab:emn_results}
    \vspace{2pt}
    \resizebox{\linewidth}{!}{
    \begin{tabular}{lcccc}
    \toprule
    \textbf{Method} & \textbf{SSR $\uparrow$} & \textbf{SSPL $\uparrow$} & \textbf{SR $\uparrow$} & \textbf{SPL $\uparrow$} \\
    \midrule
    \multicolumn{5}{l}{\textit{GOAT-Bench Baselines}} \\
    \midrule
    Modular GOAT & 0.0 & 0.0 & 18.0 & 10.6 \\
    SenseAct-NN  & 0.0 & 0.0 & 29.3 & 24.5 \\
    \midrule
    \multicolumn{5}{l}{\textit{Open-Sourced Qwen2.5VL Exploration}} \\
    \midrule
    Explore-Mem   & 1.7   &  0.4 & 38.7 & 32.0\\ 
    GraphEQA & 1.7 & 0.8 & 35.0 & 20.3\\
    3D-Mem         & 5.0  & 4.7 & 52.3 & 41.6 \\
    PredEQA  & 1.7 & 0.8 & 38.3 & 22.2 \\
    3DSPMR w/o Geo & 5.0 & 3.3  & 47.3 & 42.7 \\
    \textbf{3DSPMR (Ours)}  & \textbf{25.0} & \textbf{15.5} & \textbf{68.0} & \textbf{59.2} \\
    \bottomrule
    \end{tabular}
    }
    \vspace{-10pt} 
\end{wraptable}
In contrast, the superior SR and SPL scores demonstrate that 3DSPMR is more resilient and efficient in challenging environments where infeasible queries are present.
Furthermore, the significant gains in SSR and SSPL underscore the 3DSPMR's robust capability for spatial knowledge reuse. 
By leveraging a unified spatial memory alongside FoV-guided reasoning, 3DSPMR seamlessly retrieves prior observations and minimizes redundant exploration.
As a result, our approach more effectively addresses the fundamental challenges of long-horizon, sequential embodied tasks.

\begin{wraptable}{r}{0.4\textwidth}
\centering
\vspace{-12pt}
\caption{
Single-Task Experiments on HM-EQA.
}
\vspace{-5pt}
\begin{tabular}{lcccc}
\toprule
\textbf{Method} & \textbf{SR $\uparrow$} & \textbf{SPL $\uparrow$} \\
\midrule
Explore-Mem & 30.6  & 24.5 \\
3D-Mem & 30.8  & 25.6 \\
\textbf{3DSPMR (Ours)} & \textbf{35.8} & \textbf{27.9} \\
\bottomrule
\end{tabular}
\label{tab:eqa_results_hm_single}
\end{wraptable}
\textbf{Single Task.}
Furthermore, we evaluated 3DSPMR in the standard single-task scenario to verify its performance. 
Notably, this traditional setting can be viewed as a degenerated version of our proposed sequential setting. 
Our setting involves multiple tasks per episode with a mix of feasible and infeasible queries, while the traditional setting is limited to a single, always-feasible task per episode.

Specifically, we conducted EQA experiments on the HM-EQA dataset \cite{ren2024explore}. 
We compared 3DSPMR with two active exploration baselines, Explore-Mem \cite{ren2024explore} and 3D-Mem \cite{3dmem}, using Qwen2.5VL-72B \cite{bai2025qwen2} as the backbone.
Following established evaluation protocols \cite{majumdar2024openeqa}, we reported SR and SPL.
As shown in Table \ref{tab:eqa_results_hm_single}, 3DSPMR consistently outperforms all baselines even in these simplified single-task cases. 
These results highlight that our method excels not only in complex sequential scenarios but also generalizes robustly to standard paradigms.

\begin{wraptable}{r}{0.62\textwidth}
\centering
\vspace{-13pt}
\caption{
We conducted a detailed ablation study on EMN track. (1) w/o FoV-Reason, (2) w/o FoV-Explore, (3) w/o Geo (with neither FoV-Reason nor FoV-Explore), and (4) the full 3DSPMR, which integrates both modules and achieves peak performance across all metrics. Bold indicates the best performance.
}
\resizebox{\linewidth}{!}{
\begin{tabular}{lcccc}
\toprule
\textbf{Method} & \textbf{SSR $\uparrow$} & \textbf{SSPL $\uparrow$} & \textbf{SR $\uparrow$} & \textbf{SPL $\uparrow$} \\
\midrule
3DSPMR w/o FoV-Reason   & 8.3 & 5.2 & 56.3 & 49.6\\
3DSPMR w/o FoV-Explore & 11.7 & 9.2 & 53.7 & 43.0\\
3DSPMR w/o Geo         & 5.0 & 3.3 & 47.3 & 42.7 \\
\textbf{3DSPMR}        & \textbf{25.0} & \textbf{15.5} & \textbf{68.0} & \textbf{59.2} \\
\bottomrule
\end{tabular}}
\label{tab:ablation_detailed}
\end{wraptable}

\textbf{Ablation Study.}
To investigate the individual and synergistic contributions of our geometric components, we conducted a fine-grained ablation study on the SEER-Bench EMN track. 
The ``Geo'' mechanism represents the unified integration of both FoV-Reasoning (grounded geometric examination) and FoV-Exploration (geometric heuristics). 
By systematically removing these modules, we quantified their impacts on the agent's ability to reuse spatial knowledge and detect task infeasibility in the long-horizon sequential setting. 
The experimental results are shown in Table~\ref{tab:ablation_detailed}.

Notably, the ``w/o Geo'' method, which removes both FoV-based reasoning and exploration, shows the most substantial performance drop across every metric.
Disabling the geometric examination mechanism in reasoning results in an SSR of 8.3\% (a decrease of 16.7 percentage points).
Removing the geometric exploration heuristics drops the SSPL from 15.5\% to 9.2\%. 
This loss leads to repeated exploration, a behavior where the agent lacks the geometric incentives to prioritize underexplored regions and instead moves redundantly within previously explored space.
This validates that explicit geometric modeling is essential for maintaining spatial consistency and enabling efficient knowledge reuse across a continuous sequence of tasks.
With both geometric reasoning and exploration, our full 3DSPMR method overcomes both premature termination and repeated exploration problems, therefore showing the best overall performance. 
Detailed analyses regarding hyperparameter settings are presented in the Appendix.

\begin{table*}[t]
\centering
\caption{SEER-Bench EQA Results over Long Horizons.}
\label{tab:seer-eqa10-12env}
\resizebox{\textwidth}{!}{%
\begin{tabular}{lcccc cccc cccc}
\toprule
 & \multicolumn{4}{c}{GPT-5} & \multicolumn{4}{c}{Gemini-3.6-Flash} & \multicolumn{4}{c}{Llama-3.3-70B} \\
\cmidrule(lr){2-5}\cmidrule(lr){6-9}\cmidrule(lr){10-13}
\textbf{Method} & \textbf{SSR} $\uparrow$ & \textbf{SSPL} $\uparrow$ & \textbf{SR} $\uparrow$ & \textbf{SPL} $\uparrow$ & \textbf{SSR} $\uparrow$ & \textbf{SSPL} $\uparrow$ & \textbf{SR} $\uparrow$ & \textbf{SPL} $\uparrow$ & \textbf{SSR} $\uparrow$ & \textbf{SSPL} $\uparrow$ & \textbf{SR} $\uparrow$ & \textbf{SPL} $\uparrow$ \\
\midrule
\multicolumn{13}{l}{\textit{Blind LLMs}} \\
\midrule
Qwen3-14B & 0.0 & $\backslash$ & 28.5 & $\backslash$ & 0.0 & $\backslash$ & 27.9 & $\backslash$ & 0.0 & $\backslash$ & 30.6 & $\backslash$ \\
GPT-5 & 0.0 & $\backslash$ & 23.1 & $\backslash$ & 0.0 & $\backslash$ & 24.0 & $\backslash$ & 0.0 & $\backslash$ & 25.2 & $\backslash$ \\
\midrule
\multicolumn{13}{l}{\textit{Open-Sourced Qwen2.5VL Exploration}} \\
\midrule
Explore-Mem & 0.0 & 0.0 & 16.9 & 5.0 & 0.0 & 0.0 & 15.0 & 5.0 & 0.0 & 0.0 & 17.1 & 5.2 \\
GraphEQA & 0.0 & 0.0 & 23.5 & 12.0 & 0.0 & 0.0 & 20.4 & 11.3 & 0.0 & 0.0 & 26.2 & 13.8 \\
3D-Mem & 1.7 & 0.4 & 35.8 & 27.8 & 0.4 & 0.2 & 35.0 & 27.6 & 5.4 & 1.3 & 41.5 & 31.8 \\
PredEQA & 0.0 & 0.0 & 32.3 & 24.1 & 0.0 & 0.0 & 29.2 & 22.1 & 0.0 & 0.0 & 37.1 & 27.8 \\
\textbf{3DSPMR (Ours)} & \textbf{10.4} & \textbf{9.5} & \textbf{49.0} & \textbf{45.3} & \textbf{1.3} & \textbf{1.2} & \textbf{42.3} & \textbf{39.3} & \textbf{10.4} & \textbf{9.8} & \textbf{52.9} & \textbf{48.5} \\
\bottomrule
\end{tabular}
}
\end{table*}
\textbf{Long Horizon.}
To assess performance over a longer horizon, we conducted additional experiments using all 10 tasks per environment, rather than the 5 tasks used in the main experiments. 
We evaluated 3DSPMR and the baseline methods across 12 scenarios in the EQA track, adhering to the same evaluation protocol and metrics as in the main experiments. 

\begin{wrapfigure}{r}{0.3\textwidth}
\vspace{-12pt}
  \centering
  \includegraphics
  [width=\linewidth]
  {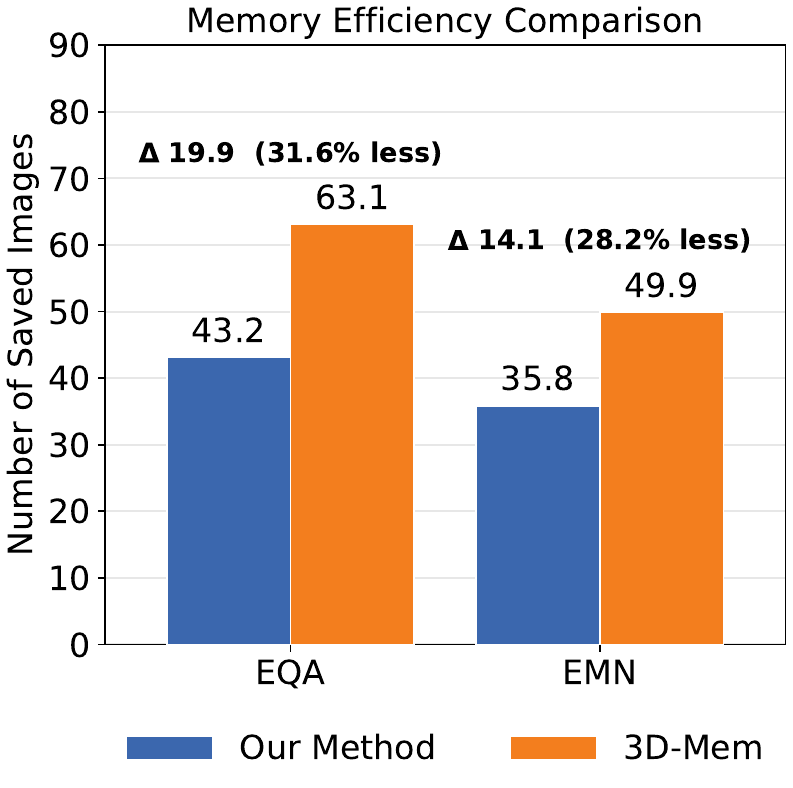
  }
  \caption{\textbf{Memory Efficiency Comparison.} On both tracks of SEER-Bench, 3DSPMR saves significantly fewer images than 3D-Mem in the memory.
  }
\label{fig:memory_comparison}
\end{wrapfigure}

As shown in Table~\ref{tab:seer-eqa10-12env}, the results reveal two main observations.
First, even under the more challenging setting with 10 tasks per environment, 3DSPMR still consistently outperforms all baseline methods across all evaluation settings.
Second, in several evaluation cases, we observe that the gaps between the correctness metrics and their corresponding efficiency metrics become smaller as the task sequence grows longer, i.e., the gaps between SR and SPL and between SSR and SSPL.
For example, while the gap between SR and SPL nearly reaches 10 percentage points under the 5-task setup, it generally stays within 5 percentage points under the 10-task setup.
This trend suggests that memory reuse becomes more beneficial over longer task sequences. 
By retaining spatial information acquired from earlier tasks, 3DSPMR can answer subsequent queries and plan exploration with less repeated exploration of previously visited regions. 

\textbf{Memory Efficiency.} To validate the compactness of the unified spatial memory in 3DSPMR, we compared the memory efficiency with 3D-Mem~\cite{3dmem}.
Specifically, we averaged the number of saved images in each episode for experiments on SEER-Bench.
As shown in Fig.~\ref{fig:memory_comparison}, 3DSPMR reduces image memory usage by 31.6\% on EQA and 28.2\% on EMN. 
3DSPMR stores substantially fewer images than 3D-Mem which retains observations based on object-clustering. 
The reduction arises from our FoV-driven gating mechanism, which filters redundant observations while retaining essential visual information. 
This is an important capability for avoiding memory saturation in long-horizon sequential tasks.
Therefore, beyond reasoning and exploration improvements, our 3DSPMR features a highly efficient and compact memory system. 
\section{Conclusion}
\label{sec:conclusion} 
We propose 3DSPMR, an approach that augments MLLMs with a unified 3D spatial memory integrating explicit geometric FoV coverage, to address challenges in sequential embodied tasks. 
This geometric grounding effectively enhances spatial reasoning and optimizes exploration efficiency. 
Additionally, we introduce SEER-Bench, a rigorous framework to comprehensively evaluate embodied agents on sequential, feasibility-aware tasks. 
Experiments demonstrate that 3DSPMR achieves state-of-the-art performance in sequential embodied tasks, delivering superior correctness and efficiency.
Future work will focus on extending this approach to dynamic environments and facilitating sim-to-real transfer.
\clearpage
\section{Acknowledgement}
This research was partially supported by the Army Research Office (ARO) grant W911NF-24-1-0385, the Office of Naval Research (ONR) grant N00014-23-1-2417, and DARPA under Award No. HR00112490426. The views, opinions, findings, and conclusions or recommendations expressed in this material are those of the authors and should not be interpreted as representing the official policies or views, either expressed or implied, of ARO, ONR, or DARPA.
\bibliographystyle{unsrt}
\bibliography{main}

@String(CVPR  = {IEEE Conf. Comput. Vis. Pattern Recog.})

@String(AAAI  = {AAAI})

@String(CVPR  = {CVPR})

@String(CVPR= {IEEE Conf. Comput. Vis. Pattern Recog.})

@String(AAAI = {AAAI})

@inproceedings{3dmem,
  title={3D-mem: 3D scene memory for embodied exploration and reasoning},
  author={Yang, Yuncong and Yang, Han and Zhou, Jiachen and Chen, Peihao and Zhang, Hongxin and Du, Yilun and Gan, Chuang},
  booktitle={Proceedings of the Computer Vision and Pattern Recognition Conference},
  pages={17294--17303},
  year={2025}
}

@article{sgnav,
  title={Sg-nav: Online 3d scene graph prompting for llm-based zero-shot object navigation},
  author={Yin, Hang and Xu, Xiuwei and Wu, Zhenyu and Zhou, Jie and Lu, Jiwen},
  journal={Advances in neural information processing systems},
  volume={37},
  pages={5285--5307},
  year={2024}
}

@inproceedings{das2018embodied,
  title={Embodied question answering},
  author={Das, Abhishek and Datta, Samyak and Gkioxari, Georgia and Lee, Stefan and Parikh, Devi and Batra, Dhruv},
  booktitle={Proceedings of the IEEE conference on computer vision and pattern recognition},
  pages={1--10},
  year={2018}
}

@inproceedings{gu2024conceptgraphs,
  title={Conceptgraphs: Open-vocabulary 3d scene graphs for perception and planning},
  author={Gu, Qiao and Kuwajerwala, Ali and Morin, Sacha and Jatavallabhula, Krishna Murthy and Sen, Bipasha and Agarwal, Aditya and Rivera, Corban and Paul, William and Ellis, Kirsty and Chellappa, Rama and others},
  booktitle={2024 IEEE International Conference on Robotics and Automation (ICRA)},
  pages={5021--5028},
  year={2024},
  organization={IEEE}
}

@article{hurst2024gpt4o,
  title   = {GPT-4o System Card},
  author  = {Hurst, Aaron and others},
  journal = {arXiv preprint arXiv:2410.21276},
  year    = {2024},
  url     = {https://arxiv.org/abs/2410.21276}
}

@article{wang2024qwen2vl,
  title={Qwen2-vl: Enhancing vision-language model's perception of the world at any resolution},
  author={Wang, Peng and Bai, Shuai and Tan, Sinan and Wang, Shijie and Fan, Zhihao and Bai, Jinze and Chen, Keqin and Liu, Xuejing and Wang, Jialin and Ge, Wenbin and others},
  journal={arXiv preprint arXiv:2409.12191},
  year={2024}
}

@article{li2024llavaonevision,
  title={Llava-onevision: Easy visual task transfer},
  author={Li, Bo and Zhang, Yuanhan and Guo, Dong and Zhang, Renrui and Li, Feng and Zhang, Hao and Zhang, Kaichen and Zhang, Peiyuan and Li, Yanwei and Liu, Ziwei and others},
  journal={arXiv preprint arXiv:2408.03326},
  year={2024}
}

@article{mu2023embodiedgpt,
  title   = {EmbodiedGPT: Vision-Language Pre-Training via Embodied Chain of Thought},
  author  = {Mu, Yao and Zhang, Qinglong and Hu, Mengkang and Wang, Wenhai and Ding, Mingyu and Jin, Jun and Wang, Bin and Dai, Jifeng and Qiao, Yu and Luo, Ping},
  journal = {arXiv preprint arXiv:2305.15021},
  year    = {2023},
  url     = {https://arxiv.org/abs/2305.15021}
}

@InProceedings{goetting2025vlmnav,
  title = 	 {End-to-End Navigation with Vision-Language Models: Transforming Spatial Reasoning into Question-Answering},
  author =       {Goetting, Dylan and Singh, Himanshu Gaurav and Loquercio, Antonio},
  booktitle = 	 {Proceedings of the International Conference on Neuro-symbolic Systems},
  pages = 	 {22--35},
  year = 	 {2025},
  volume = 	 {288},
  series = 	 {Proceedings of Machine Learning Research},
  month = 	 {28--30 May},
  publisher =    {PMLR},
  url = 	 {https://proceedings.mlr.press/v288/goetting25a.html},
}

@inproceedings{liu2024llavaplus,
  title={Llava-plus: Learning to use tools for creating multimodal agents},
  author={Liu, Shilong and Cheng, Hao and Liu, Haotian and Zhang, Hao and Li, Feng and Ren, Tianhe and Zou, Xueyan and Yang, Jianwei and Su, Hang and Zhu, Jun and others},
  booktitle={European conference on computer vision},
  pages={126--142},
  year={2024},
  organization={Springer}
}

@inproceedings{werby2024hovsg,
  title     = {Hierarchical Open-Vocabulary 3D Scene Graphs for Language-Grounded Robot Navigation},
  author    = {Werby, Abdelrhman and Huang, Chenguang and B{\"u}chner, Martin and Valada, Abhinav and Burgard, Wolfram},
  booktitle = {Robotics: Science and Systems (RSS)},
  year      = {2024},
  url       = {https://www.roboticsproceedings.org/rss20/p077.pdf}
}

@inproceedings{gordon2018iqa,
  title     = {IQA: Visual Question Answering in Interactive Environments},
  author    = {Gordon, Daniel and Kembhavi, Aniruddha and Rastegari, Mohammad and Redmon, Joseph and Fox, Dieter and Farhadi, Ali},
  booktitle = {Proceedings of the IEEE/CVF Conference on Computer Vision and Pattern Recognition (CVPR)},
  year      = {2018},
  url       = {https://openaccess.thecvf.com/content_cvpr_2018/papers/Gordon_IQA_Visual_Question_CVPR_2018_paper.pdf}
}

@inproceedings{yang2024emma,
  title     = {Embodied Multi-Modal Agent trained by an LLM from a Parallel TextWorld},
  author    = {Yang, Yijun and Zhou, Tianyi and Li, Kanxue and Tao, Dapeng and Li, Lusong and Shen, Li and He, Xiaodong and Jiang, Jing and Shi, Yuhui},
  booktitle = {Proceedings of the IEEE/CVF Conference on Computer Vision and Pattern Recognition (CVPR)},
  year      = {2024},
  url       = {https://openaccess.thecvf.com/content/CVPR2024/papers/Yang_Embodied_Multi-Modal_Agent_trained_by_an_LLM_from_a_Parallel_CVPR_2024_paper.pdf}
}

@article{saxena2024grapheqa,
  title={Grapheqa: Using 3d semantic scene graphs for real-time embodied question answering},
  author={Saxena, Saumya and Buchanan, Blake and Paxton, Chris and Liu, Peiqi and Chen, Bingqing and Vaskevicius, Narunas and Palmieri, Luigi and Francis, Jonathan and Kroemer, Oliver},
  journal={arXiv preprint arXiv:2412.14480},
  year={2024}
}

@inproceedings{majumdar2024openeqa,
  title={Openeqa: Embodied question answering in the era of foundation models},
  author={Majumdar, Arjun and Ajay, Anurag and Zhang, Xiaohan and Putta, Pranav and Yenamandra, Sriram and Henaff, Mikael and Silwal, Sneha and Mcvay, Paul and Maksymets, Oleksandr and Arnaud, Sergio and others},
  booktitle={2024 IEEE/CVF Conference on Computer Vision and Pattern Recognition (CVPR)},
  pages={16488--16498},
  year={2024},
  organization={IEEE}
}

@article{zhao2025cityeqa,
  title   = {CityEQA: A Hierarchical LLM Agent on Embodied Question Answering Benchmark in City Space},
  author  = {Zhao, Yong and Xu, Kai and Zhu, Zhengqiu and Hu, Yue and Zheng, Zhiheng and Chen, Yingfeng and Ji, Yatai and Gao, Chen and Li, Yong and Huang, Jincai},
  journal = {arXiv preprint arXiv:2502.12532},
  year    = {2025},
  url     = {https://arxiv.org/abs/2502.12532}
}

@article{jiang2025explorationeqa,
  title   = {Beyond the Destination: A Novel Benchmark for Exploration-Aware Embodied Question Answering},
  author  = {Jiang, Kaixuan and Liu, Yang and Chen, Weixing and Luo, Jingzhou and Chen, Ziliang and Pan, Ling and Li, Guanbin and Lin, Liang},
  journal = {arXiv preprint arXiv:2503.11117},
  year    = {2025},
  url     = {https://arxiv.org/abs/2503.11117}
}

@inproceedings{zheng2024navillm,
  title     = {Towards Learning a Generalist Model for Embodied Navigation},
  author    = {Zheng, Duo and Huang, Shijia and Zhao, Lin and Zhong, Yiwu and Wang, Liwei},
  booktitle = {Proceedings of the IEEE/CVF Conference on Computer Vision and Pattern Recognition (CVPR)},
  year      = {2024},
  url       = {https://openaccess.thecvf.com/content/CVPR2024/papers/Zheng_Towards_Learning_a_Generalist_Model_for_Embodied_Navigation_CVPR_2024_paper.pdf}
}

@inproceedings{song2025lhvln,
  title={Towards long-horizon vision-language navigation: Platform, benchmark and method},
  author={Song, Xinshuai and Chen, Weixing and Liu, Yang and Chen, Weikai and Li, Guanbin and Lin, Liang},
  booktitle={2025 IEEE/CVF Conference on Computer Vision and Pattern Recognition (CVPR)},
  pages={12078--12088},
  year={2025},
  organization={IEEE}
}

@inproceedings{yin2025unigoal,
  title     = {UniGoal: Towards Universal Zero-shot Goal-oriented Navigation},
  author    = {Yin, Hang and Xu, Xiuwei and Zhao, Linqing and Wang, Ziwei and Zhou, Jie and Lu, Jiwen},
  booktitle = {Proceedings of the IEEE/CVF Conference on Computer Vision and Pattern Recognition (CVPR)},
  year      = {2025},
  url       = {https://openaccess.thecvf.com/content/CVPR2025/papers/Yin_UniGoal_Towards_Universal_Zero-shot_Goal-oriented_Navigation_CVPR_2025_paper.pdf}
}

@inproceedings{liu2024ver,
  title     = {Volumetric Environment Representation for Vision-Language Navigation},
  author    = {Liu, R. and others},
  booktitle = {Proceedings of the IEEE/CVF Conference on Computer Vision and Pattern Recognition (CVPR)},
  year      = {2024},
  url       = {https://openaccess.thecvf.com/content/CVPR2024/html/Liu_Volumetric_Environment_Representation_for_Vision-Language_Navigation_CVPR_2024_paper.html}
}

@inproceedings{yadav2023habitat,
  title={Habitat-matterport 3d semantics dataset},
  author={Yadav, Karmesh and Ramrakhya, Ram and Ramakrishnan, Santhosh Kumar and Gervet, Theo and Turner, John and Gokaslan, Aaron and Maestre, Noah and Chang, Angel Xuan and Batra, Dhruv and Savva, Manolis and others},
  booktitle={Proceedings of the IEEE/CVF Conference on Computer Vision and Pattern Recognition},
  pages={4927--4936},
  year={2023}
}

@article{szot2021habitat,
  title={Habitat 2.0: Training home assistants to rearrange their habitat},
  author={Szot, Andrew and Clegg, Alexander and Undersander, Eric and Wijmans, Erik and Zhao, Yili and Turner, John and Maestre, Noah and Mukadam, Mustafa and Chaplot, Devendra Singh and Maksymets, Oleksandr and others},
  journal={Advances in neural information processing systems},
  volume={34},
  pages={251--266},
  year={2021}
}

@inproceedings{khanna2024goat,
  title={Goat-bench: A benchmark for multi-modal lifelong navigation},
  author={Khanna, Mukul and Ramrakhya, Ram and Chhablani, Gunjan and Yenamandra, Sriram and Gervet, Theophile and Chang, Matthew and Kira, Zsolt and Chaplot, Devendra Singh and Batra, Dhruv and Mottaghi, Roozbeh},
  booktitle={Proceedings of the IEEE/CVF Conference on Computer Vision and Pattern Recognition},
  pages={16373--16383},
  year={2024}
}

@inproceedings{yang2025thinking,
  title={Thinking in space: How multimodal large language models see, remember, and recall spaces},
  author={Yang, Jihan and Yang, Shusheng and Gupta, Anjali W and Han, Rilyn and Fei-Fei, Li and Xie, Saining},
  booktitle={Proceedings of the Computer Vision and Pattern Recognition Conference},
  pages={10632--10643},
  year={2025}
}

@article{chen2025spatial,
  title={Why is spatial reasoning hard for vlms? an attention mechanism perspective on focus areas},
  author={Chen, Shiqi and Zhu, Tongyao and Zhou, Ruochen and Zhang, Jinghan and Gao, Siyang and Niebles, Juan Carlos and Geva, Mor and He, Junxian and Wu, Jiajun and Li, Manling},
  journal={arXiv preprint arXiv:2503.01773},
  year={2025}
}

@inproceedings{wijmans2019embodied,
  title={Embodied question answering in photorealistic environments with point cloud perception},
  author={Wijmans, Erik and Datta, Samyak and Maksymets, Oleksandr and Das, Abhishek and Gkioxari, Georgia and Lee, Stefan and Essa, Irfan and Parikh, Devi and Batra, Dhruv},
  booktitle={Proceedings of the IEEE/CVF Conference on Computer Vision and Pattern Recognition},
  pages={6659--6668},
  year={2019}
}

@inproceedings{zhou2024navgpt,
  title={Navgpt: Explicit reasoning in vision-and-language navigation with large language models},
  author={Zhou, Gengze and Hong, Yicong and Wu, Qi},
  booktitle={Proceedings of the AAAI Conference on Artificial Intelligence},
  volume={38},
  number={7},
  pages={7641--7649},
  year={2024}
}

@inproceedings{yu2019multi,
  title={Multi-target embodied question answering},
  author={Yu, Licheng and Chen, Xinlei and Gkioxari, Georgia and Bansal, Mohit and Berg, Tamara L and Batra, Dhruv},
  booktitle={Proceedings of the IEEE/CVF Conference on Computer Vision and Pattern Recognition},
  pages={6309--6318},
  year={2019}
}

@article{batra2020objectnav,
  title={Objectnav revisited: On evaluation of embodied agents navigating to objects},
  author={Batra, Dhruv and Gokaslan, Aaron and Kembhavi, Aniruddha and Maksymets, Oleksandr and Mottaghi, Roozbeh and Savva, Manolis and Toshev, Alexander and Wijmans, Erik},
  journal={arXiv preprint arXiv:2006.13171},
  year={2020}
}

@inproceedings{zhang20233d,
  title={3d-aware object goal navigation via simultaneous exploration and identification},
  author={Zhang, Jiazhao and Dai, Liu and Meng, Fanpeng and Fan, Qingnan and Chen, Xuelin and Xu, Kai and Wang, He},
  booktitle={Proceedings of the IEEE/CVF Conference on Computer Vision and Pattern Recognition},
  pages={6672--6682},
  year={2023}
}

@article{ren2024explore,
  title={Explore until confident: Efficient exploration for embodied question answering},
  author={Ren, Allen Z and Clark, Jaden and Dixit, Anushri and Itkina, Masha and Majumdar, Anirudha and Sadigh, Dorsa},
  journal={arXiv preprint arXiv:2403.15941},
  year={2024}
}

@inproceedings{deitke2020robothor,
  title={Robothor: An open simulation-to-real embodied ai platform},
  author={Deitke, Matt and Han, Winson and Herrasti, Alvaro and Kembhavi, Aniruddha and Kolve, Eric and Mottaghi, Roozbeh and Salvador, Jordi and Schwenk, Dustin and VanderBilt, Eli and Wallingford, Matthew and others},
  booktitle={Proceedings of the IEEE/CVF conference on computer vision and pattern recognition},
  pages={3164--3174},
  year={2020}
}

@article{du2025vl,
  title={VL-Nav: Real-time Vision-Language Navigation with Spatial Reasoning},
  author={Du, Yi and Fu, Taimeng and Chen, Zhuoqun and Li, Bowen and Su, Shaoshu and Zhao, Zhipeng and Wang, Chen},
  journal={arXiv preprint arXiv:2502.00931},
  year={2025}
}

@misc{openai2025gpt5,
  title         = {GPT-5 Technical Overview},
  author        = {OpenAI},
  howpublished  = {\url{https://openai.com/index/introducing-gpt-5/}},
  year          = {2025}
}

@misc{deepmind2026gemini36flash,
  title={Gemini 3.6 Flash Model Card},
  author={{Google DeepMind}},
  year={2026},
  howpublished={\url{https://deepmind.google/models/model-cards/gemini-3-6-flash/}}
}

@misc{meta2024llama33,
  author       = {{Meta}},
  title        = {{Llama 3.3 70B Instruct}},
  year         = {2024},
  howpublished = {\url{https://huggingface.co/meta-llama/Llama-3.3-70B-Instruct}}
}

@article{yang2025qwen3,
  title={Qwen3 technical report},
  author={Yang, An and Li, Anfeng and Yang, Baosong and Zhang, Beichen and Hui, Binyuan and Zheng, Bo and Yu, Bowen and Gao, Chang and Huang, Chengen and Lv, Chenxu and others},
  journal={arXiv preprint arXiv:2505.09388},
  year={2025}
}

@article{bai2025qwen2,
  title={Qwen2.5-vl technical report},
  author={Bai, Shuai and Chen, Keqin and Liu, Xuejing and Wang, Jialin and Ge, Wenbin and Song, Sibo and Dang, Kai and Wang, Peng and Wang, Shijie and Tang, Jun and others},
  journal={arXiv preprint arXiv:2502.13923},
  year={2025}
}

@article{comanici2025gemini,
  title={Gemini 2.5: Pushing the frontier with advanced reasoning, multimodality, long context, and next generation agentic capabilities},
  author={Comanici, Gheorghe and Bieber, Eric and Schaekermann, Mike and Pasupat, Ice and Sachdeva, Noveen and Dhillon, Inderjit and Blistein, Marcel and Ram, Ori and Zhang, Dan and Rosen, Evan and others},
  journal={arXiv preprint arXiv:2507.06261},
  year={2025}
}

@article{hughes2022hydra,
  title={Hydra: A real-time spatial perception system for 3D scene graph construction and optimization},
  author={Hughes, Nathan and Chang, Yun and Carlone, Luca},
  journal={arXiv preprint arXiv:2201.13360},
  year={2022}
}

@article{hughes2024foundations,
  title={Foundations of spatial perception for robotics: Hierarchical representations and real-time systems},
  author={Hughes, Nathan and Chang, Yun and Hu, Siyi and Talak, Rajat and Abdulhai, Rumaia and Strader, Jared and Carlone, Luca},
  journal={The International Journal of Robotics Research},
  volume={43},
  number={10},
  pages={1457--1505},
  year={2024},
  publisher={SAGE Publications Sage UK: London, England}
}

@article{chang2023goat,
  title={Goat: Go to any thing},
  author={Chang, Matthew and Gervet, Theophile and Khanna, Mukul and Yenamandra, Sriram and Shah, Dhruv and Min, So Yeon and Shah, Kavit and Paxton, Chris and Gupta, Saurabh and Batra, Dhruv and others},
  journal={arXiv preprint arXiv:2311.06430},
  year={2023}
}

@article{rana2023sayplan,
  title={Sayplan: Grounding large language models using 3d scene graphs for scalable robot task planning},
  author={Rana, Krishan and Haviland, Jesse and Garg, Sourav and Abou-Chakra, Jad and Reid, Ian and Suenderhauf, Niko},
  journal={arXiv preprint arXiv:2307.06135},
  year={2023}
}

@inproceedings{yokoyama2024hm3d,
  title={Hm3d-ovon: A dataset and benchmark for open-vocabulary object goal navigation},
  author={Yokoyama, Naoki and Ramrakhya, Ram and Das, Abhishek and Batra, Dhruv and Ha, Sehoon},
  booktitle={2024 IEEE/RSJ International Conference on Intelligent Robots and Systems (IROS)},
  pages={5543--5550},
  year={2024},
  organization={IEEE}
}

@article{long2024instructnav,
  title={Instructnav: Zero-shot system for generic instruction navigation in unexplored environment},
  author={Long, Yuxing and Cai, Wenzhe and Wang, Hongcheng and Zhan, Guanqi and Dong, Hao},
  journal={arXiv preprint arXiv:2406.04882},
  year={2024}
}

@inproceedings{dang2025ecbench,
  title={Ecbench: Can multi-modal foundation models understand the egocentric world? a holistic embodied cognition benchmark},
  author={Dang, Ronghao and Yuan, Yuqian and Zhang, Wenqi and Xin, Yifei and Zhang, Boqiang and Li, Long and Wang, Liuyi and Zeng, Qinyang and Li, Xin and Bing, Lidong},
  booktitle={Proceedings of the Computer Vision and Pattern Recognition Conference},
  pages={24593--24602},
  year={2025}
}

@article{linghu2024multi,
  title={Multi-modal situated reasoning in 3d scenes},
  author={Linghu, Xiongkun and Huang, Jiangyong and Niu, Xuesong and Ma, Xiaojian Shawn and Jia, Baoxiong and Huang, Siyuan},
  journal={Advances in Neural Information Processing Systems},
  volume={37},
  pages={140903--140936},
  year={2024}
}

@inproceedings{kirillov2023segment,
  title={Segment anything},
  author={Kirillov, Alexander and Mintun, Eric and Ravi, Nikhila and Mao, Hanzi and Rolland, Chloe and Gustafson, Laura and Xiao, Tete and Whitehead, Spencer and Berg, Alexander C and Lo, Wan-Yen and others},
  booktitle={Proceedings of the IEEE/CVF international conference on computer vision},
  pages={4015--4026},
  year={2023}
}

@inproceedings{cheng2024yolo,
  title={Yolo-world: Real-time open-vocabulary object detection},
  author={Cheng, Tianheng and Song, Lin and Ge, Yixiao and Liu, Wenyu and Wang, Xinggang and Shan, Ying},
  booktitle={Proceedings of the IEEE/CVF conference on computer vision and pattern recognition},
  pages={16901--16911},
  year={2024}
}

@article{radford2021learning,
  title={Learning Transferable Visual Models From Natural Language Supervision},
  author={Radford, Alec and Kim, Jong Wook and Hallacy, Chris and et al.},
  journal={arXiv preprint arXiv:2103.00020},
  year={2021}
}

@article{anderson2018evaluation,
  title={On evaluation of embodied navigation agents},
  author={Anderson, Peter and Chang, Angel and Chaplot, Devendra Singh and Dosovitskiy, Alexey and Gupta, Saurabh and Koltun, Vladlen and Kosecka, Jana and Malik, Jitendra and Mottaghi, Roozbeh and Savva, Manolis and others},
  journal={arXiv preprint arXiv:1807.06757},
  year={2018}
}

@article{wang2025think,
  title={Aux-think: Exploring reasoning strategies for data-efficient vision-language navigation},
  author={Wang, Shuo and Wang, Yongcai and Li, Wanting and Cai, Xudong and Wang, Yucheng and Chen, Maiyue and Wang, Kaihui and Su, Zhizhong and Li, Deying and Fan, Zhaoxin},
  journal={arXiv preprint arXiv:2505.11886},
  year={2025}
}

@article{li2025industryeqa,
  title={IndustryEQA: Pushing the Frontiers of Embodied Question Answering in Industrial Scenarios},
  author={Li, Yifan and Chen, Yuhang and Dao, Anh and Li, Lichi and Cai, Zhongyi and Tan, Zhen and Chen, Tianlong and Kong, Yu},
  journal={arXiv preprint arXiv:2505.20640},
  year={2025}
}

@inproceedings{yuan2026predict,
  title={Predict Before You Explore: Predictive Planning with Specialized Memory for Embodied Question Answering},
  author={Yuan, Bowen and You, Sisi and Bao, Bing-Kun},
  booktitle={Proceedings of the IEEE/CVF Conference on Computer Vision and Pattern Recognition},
  pages={29610--29619},
  year={2026}
}
\clearpage
\appendix
\section{Details of SEER-Bench}
\label{sec:seer-bench_appendix}
We propose a new Sequential Embodied Exploration and Reasoning Benchmark (SEER-Bench) to evaluate agent performance on sequential embodied tasks in section~\ref{sec:seer-bench} of the main paper. 
Here, we supplement more details. 

\begin{figure}[h]
\centering
  \includegraphics
  [width=\linewidth]
  {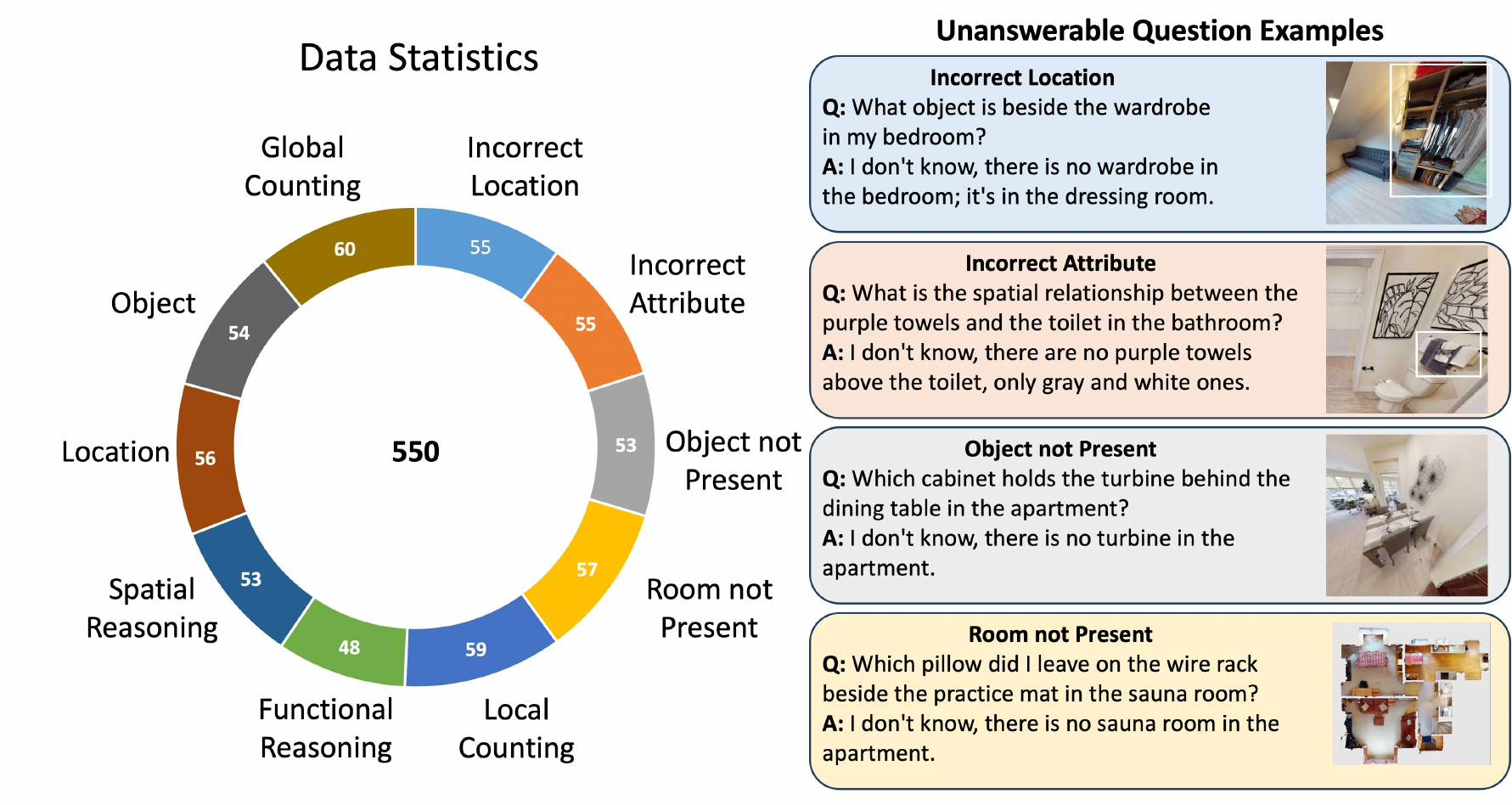}
  \caption{{Data statistics and Unanswerable Question Examples of EQA in SEER-Bench.
  } 
  }
\label{fig:benchmark}
\end{figure}
\subsection{Simulator}
HM3DSem~\cite{yadav2023habitat} is a large-scale dataset providing high-fidelity 3D reconstructions of 3,100 rooms with dense semantic annotations for objects, materials, and room structures. 
We adopted HM3DSem within the Habitat simulator~\cite{szot2021habitat} to construct photorealistic and semantically consistent embodied environments. 
Habitat supports high-performance egocentric rendering, efficient physics-based navigation, and modular interfaces for defining sensors, agents, and tasks. 
We leveraged these capabilities to create sequential task episodes that maintain consistent spatial layouts across tasks, enabling reproducible evaluation of spatial reasoning, memory reuse, and feasibility awareness in both Embodied Question Answering (EQA) and Embodied Multi-modal Navigation (EMN) settings.
In this work, we did not include cross-floor tasks, treating each floor as a separate scenario instead.

\subsection{Embodied Question Answering}
As existing HM3DSem-related benchmarks~\cite{majumdar2024openeqa, jiang2025explorationeqa} do not provide sequential question–answer pairs for each scenario, we annotated 60 HM3DSem scenarios with 5 to 10 QA pairs for each scenario.
The questions in each scenario span six normal categories: Object, Location, Spatial Reasoning, Functional Reasoning, Local Counting, and Global Counting.
In EQA, infeasible tasks naturally correspond to unanswerable questions.
To systematically capture such cases, we introduced four types of unanswerable questions: Incorrect Location (\eg, erroneous room–object associations), Incorrect Attribute (\eg, incorrect color or state of the mentioned object), Object not Present, and Room not Present.
For each scenario, we constructed a QA set with at most one question from each answerable and unanswerable category.
Data statistics and examples of unanswerable questions are provided in Fig.~\ref{fig:benchmark}.

To support data annotation, we collected environmental information from each selected scenario.
We implemented an autonomous exploration policy that, at each step, navigates the agent to the nearest frontier on the occupancy map, enabling systematic acquisition of ego-centric RGB observations.
For each environment, the agent is initialized with two random feasible starting positions and camera orientations to increase trajectory diversity and improve spatial coverage.

Using these collected frames, we employed GPT-5~\cite{openai2025gpt5} to generate most of the QA annotations, followed by manual filtering and correction to ensure correctness.
For global counting questions, we obtained ground-truth answers by manually inspecting the layout figures or referencing the metadata.
For unanswerable questions, GPT-5 is used to annotate cases of Incorrect Location and Incorrect Attribute.
For the Object Not Present and Room Not Present categories, we designed two additional prompts that guide GPT-5 to generate questions involving objects (\eg, carriage) or rooms (\eg, sauna room) that are rarely found in typical residential environments.
Furthermore, we recorded some unique objects and rooms appearing in each scenario and intentionally reused them in other environments where they do not exist, enabling controlled comparisons across scenarios.

\begin{figure}[t]
\centering
  \includegraphics
  [width=\linewidth]
  {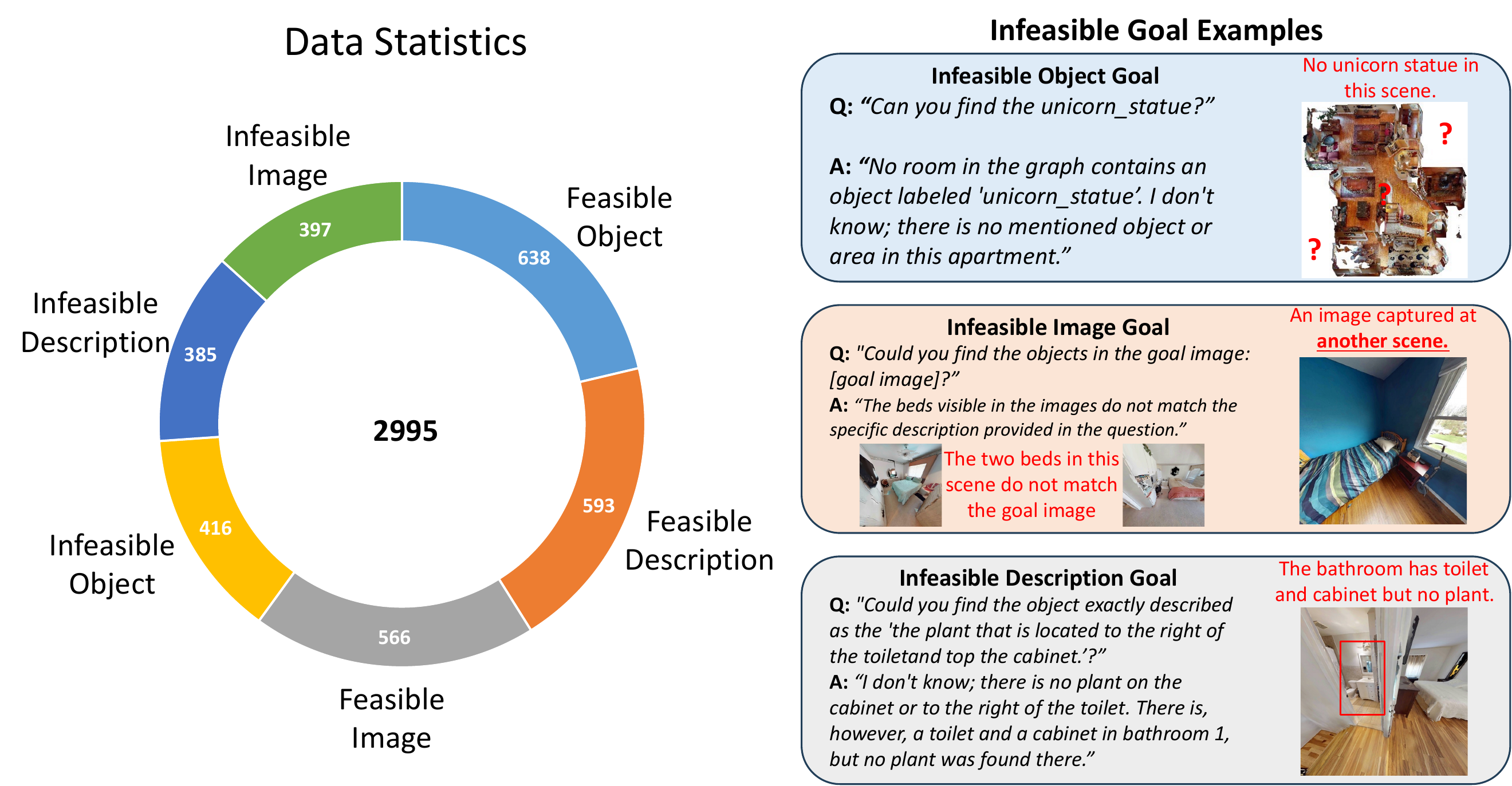}
  \caption{{Data statistics of the EMN track in SEER-Bench and infeasible goal examples.} 
  }
\label{fig:SeqEMN_track}
\end{figure}
\subsection{Embodied Multi-modal Navigation}
The Embodied Multi-modal Navigation (EMN) task in SEER builds upon the GOAT-Bench~\cite{khanna2024goat}, extending it into a sequential, feasibility-aware formulation.
Each navigation task requires an embodied agent to navigate toward a target object defined by one of three goal modalities: 
(1) an \textit{object goal}, specified by the target object category;
(2) a \textit{language goal}, described by natural language instructions; or 
(3) an \textit{image goal}, provided as a visual exemplar.
Within an episode, the agent must complete each navigation task sequentially, reusing spatial knowledge accumulated from previous tasks for efficient planning.

Each task specifies a target using one of the three modalities: \emph{object goal}, \emph{language goal}, or \emph{image goal}.  
To evaluate feasibility awareness, we introduce three corresponding types of infeasible goals:

\begin{itemize}[leftmargin=*, nosep]
    \item Object-Goal Infeasibility (Incorrect Location).  
    Following the EQA setup, we designed target object categories that do \emph{not} exist in the scene.  
    The agent must determine that the goal is impossible after sufficient exploration and terminate appropriately.

    \item Language-Goal Infeasibility (Incorrect Description).  
    We modified original GOAT-Bench language instructions by altering key attributes or spatial relations.  
    For example, an instruction such as ``the white surfboard near a plant and TV'' may be changed to  
    ``the black surfboard near a bed and oven'',  
    producing a linguistically coherent but environmentally impossible goal.

    \item Image-Goal Infeasibility (Cross-Scene Images).  
    We used an image query taken from a different HM3DSem scene than the current episode.  
    For instance, an exemplar from Scene~A is provided for a navigation task in Scene~B, making the visual reference unmatchable within the environment.
\end{itemize}
Consistent with the configuration in the EQA track, we included 60 scenarios from HM3DSem and assigned 40\% of the goals as infeasible, ensuring a balanced distribution between feasible and infeasible tasks.
As summarized in Fig.~\ref{fig:SeqEMN_track}, the EMN track contains a total of 2995 multi-modal goals.  
The right side of the figure illustrates example cases for each of the three infeasible goal types.

\subsection{Evaluation Metrics}
\label{sec:metric_appendix}
\textbf{Evaluation Metrics for EQA.}
As traditional evaluation metrics Success Rate (SR) and Success-weighted Path Length (SPL) fail to measure model performance for an episode,
we introduce two new evaluation metrics: Sequential Success Rate (SSR) for sequential correctness and Sequential SPL (SSPL) for sequential efficiency in Section~\ref{sec:seer-bench} of the main paper.
Here, we provide the SSR and SSPL formulas for Embodied Question Answering (EQA).

In an EQA task, the agent is required to infer an answer for each question. 
Consequently, the binary success indicator $S_i\in\{0,1\}$ for task $i$ in EMN is replaced by an LLM-based correctness score $\sigma_i \in \{1, \dots, 5\}$ for the inferred answer. 
In this scale, 1 indicates an incorrect response, 5 signifies a fully correct response, and intermediate values represent varying degrees of semantic similarity to the ground truth. 
We adopt the correctness and efficiency metrics defined in OpenEQA~\cite{majumdar2024openeqa} as SR and SPL.
To adapt the SSR for EQA, we consider a task successful if its score $\sigma_i$ is larger than 3. Based on this, we generalize $\text{SSR}_{\text{EMN}} = \frac{1}{|\mathcal{E}|} \sum_{e \in \mathcal{E}} \mathbb{I}[\text{SR}_e = 1]\times 100\%$ to account for the nuanced scores:
\begin{equation}
\text{SSR}_{\text{EQA}} = \frac{1}{|\mathcal{E}|} \sum_{e \in \mathcal{E}} \frac{\max(0, \sum_{i=1}^{N_e} (\sigma_i - 3))}{2 \times N_e} \times 100\%,
\end{equation}
where $\mathcal{E}$ is the set of test episodes, $e \in \mathcal{E}$ is a single episode comprising a sequence of $N_e$ tasks. 
Similarly, the SSPL for EQA is adapted from the standard EMN formulation, $\text{SSPL}_{\text{EMN}} = \frac{1}{|\mathcal{E}|} \sum_{e \in \mathcal{E}} \left( \mathbb{I}[SR_e = 1] \times \frac{L_e}{\max(H_e, L_e)} \right) \times 100\%$, by reformulating the success condition as follows:
\begin{equation}
\text{SSPL}_{\text{EQA}} = \frac{1}{|\mathcal{E}|} \sum_{e \in \mathcal{E}} \left( \frac{\max(0, \sum_{i=1}^{N_e} (\sigma_i - 3))}{2 \times N_e} \times \frac{L_e}{\max(H_e, L_e)} \right) \times 100\%.
\end{equation}

\textbf{Task Step Limit and Episode Step Limit.}
In this paragraph, we specify the task step limit, episode step limit, and step budget $L_e$ allocated for an episode $e$.
As discussed in the benchmark description, SEER-Bench includes \emph{small}, \emph{medium}, and \emph{large} indoor scenes.
For the EQA track, we set the task step limits for these three scene sizes to 50, 200, and 320 steps, respectively.
The \emph{episode step limit} is defined as twice the corresponding task step limit.
These limits approximate the number of steps required to fully explore each scene size using a closest-frontier exploration policy, ensuring adequate spatial coverage while preventing unnecessarily long trajectories.
For the EMN track, we extended the step-limit strategy used in GOAT-Bench~\cite{khanna2024goat}.
Specifically, we used Habitat's scene-loading utilities to obtain the spatial extent of each environment, assigned a size-dependent local step limit, and again set the episode limit to twice the task budget.
For both the EQA and EMN tracks, we set the expected optimal step budget $L_e$ to a specified proportion (strictly less than 1) of the episode step limit.
In both the EQA and EMN tracks, tasks that exceed the step limit are categorized as infeasible.
If the episode step limit is exceeded, the current task and all remaining tasks in the episode are marked as failed.

\begin{figure}[t]
\centering
  \includegraphics
  [width=0.96\linewidth]
  {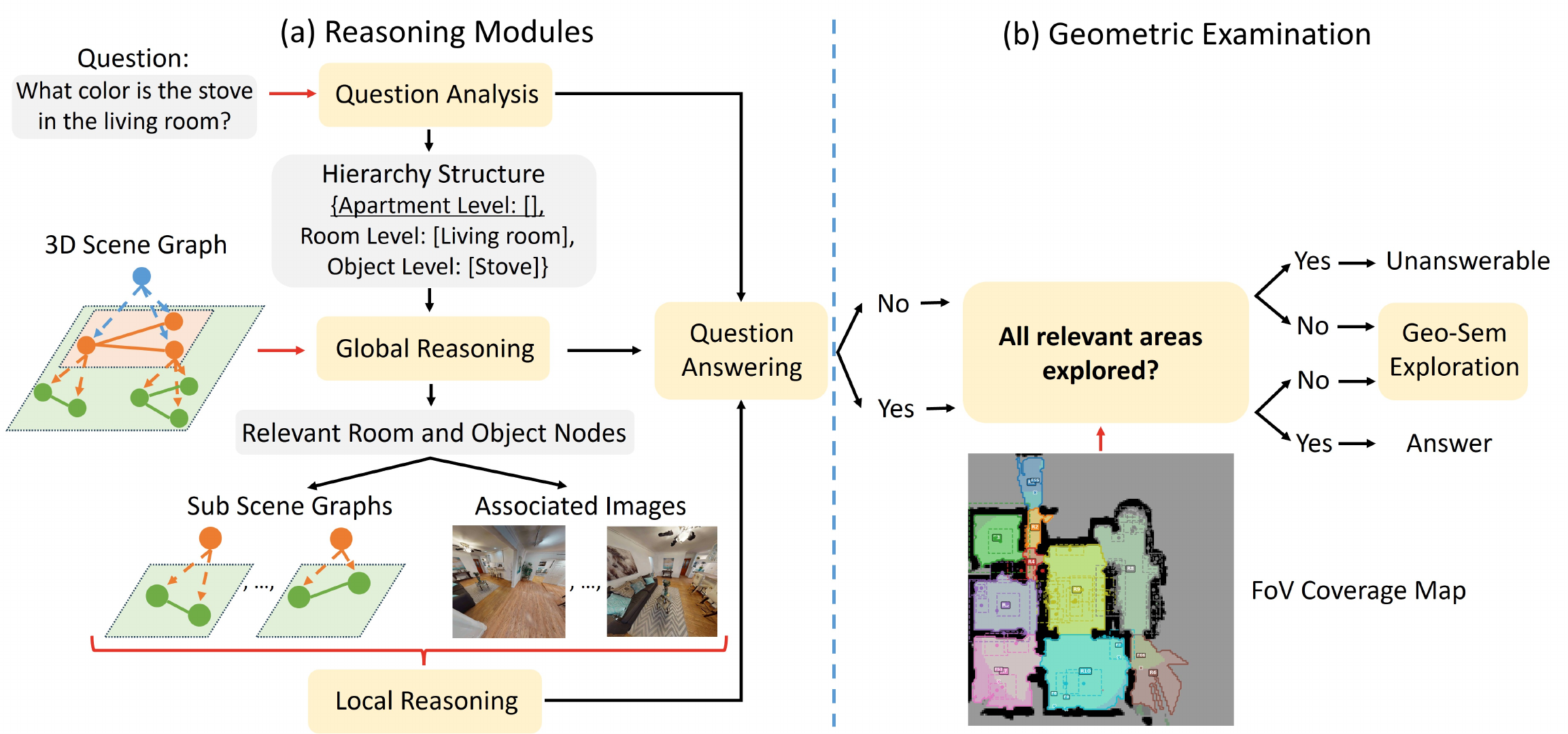}
  \caption{{\bfseries Illustration of Geo-Reasoning in 3DSPMR for EQA tasks.}
(a) Phase 1: Reasoning Modules leverage the graph and images in spatial memory to generate an initial inference.
(b) Phase 2: Geometric Examination mechanism utilizes FoV coverage to assess task completion.
  If the relevant spatial regions are fully explored, the system either confirms the inferred answer or flags task infeasibility; otherwise, it triggers Geo-Sem Exploration to determine the next optimal exploration direction.
  }
  \label{fig:agentic_reasoning}
\end{figure}

\section{Implementation}
\label{sec:implementation_appendix}
\subsection{Geo-Reasoning Details}
\label{sec:reasoning_appendix}
We introduce the Geo-Reasoning module in Section~\ref{subsec:reasoning} of the main paper.
Here, we provide a detailed description and a visual illustration of the reasoning pipeline.
As shown in Fig. \ref{fig:agentic_reasoning}, Geo-Reasoning operates in two distinct phases: first, Reasoning Modules provide an initial inference; second, a Geometric Examination Mechanism performs post-reasoning validation.
We utilize MLLMs in a training-free, hierarchical manner to interpret the multi-modal cues stored in memory. 

For the EQA task, the MLLM reasoning process is executed through four specialized stages: Question Analysis decomposes the query into hierarchical semantic requirements;
Global Reasoning prunes the 3D scene graph $G$ to isolate task-relevant sub-graphs; 
Local Reasoning then retrieves the associated keyframes $I$ indexed by these sub-graphs, further narrowing the visual search space to extract precise perceptual evidence;
and Question Answering formulates the final response by synthesizing symbolic and visual evidence. 
This pipeline effectively distills the complex, multi-modal spatial memory into task-level decisions supported by structured evidence. 
These initial conclusions are then subjected to the Geometric Examination to ensure their geometric grounding.
Depending on the extent of spatial exploration, this examination either validates or rectifies the initial outputs.

\subsection{Hyperparameters}
\noindent\textbf{FoV-Based Image Gating.} 
The unified spatial memory in 3DSPMR leverages a gating mechanism based on Field-of-View (FoV) to determine stored observation frames. 
In this mechanism, the geometric novelty score threshold $\tau_0$ controls the minimum proportion of novel spatial coverage required for an observation to be committed to the image bank. 
An observation is only stored if the calculated novelty score (representing the ratio of newly observed cells to the total cells in the current view) exceeds the default threshold of 0.25 in EQA and 0.5 in EMN.

\noindent\textbf{Geo-Sem Exploration Weights.} 
In our Geo-Sem Exploration, the agent identifies the optimal frontier target by maximizing a composite score that integrates semantic relevance, geometric exploration potential, and proximity. 
We assigned a weight $w_{\mathrm{sem}}$ of 0.5 to semantic relevance to prioritize task-pertinent visual cues, a weight $w_{\mathrm{dist}}$ of 0.4 to geometric proximity to favor closer targets, and another weight $w_{\mathrm{exp}}$ of 0.1 to unexplored area potential to promote environmental coverage. 
To prevent trajectory oscillations, the agent only switches to a new frontier target if its composite score exceeds the current target's score by a switching threshold of 0.1.

\subsection{Backbones and System Components}
We used Qwen2.5VL-72B~\cite{bai2025qwen2} as our default open-source MLLM backbone.
To construct and maintain object-level nodes in the 3D scene graph, we build upon ConceptGraphs~\cite{gu2024conceptgraphs}, which loads the following perception models:
YOLO-World \cite{cheng2024yolo} for open-vocabulary object detection,
SAM (Segment Anything Model) \cite{kirillov2023segment} for instance segmentation,
CLIP \cite{radford2021learning} for visual embedding and category grounding.
These components collectively provide reliable open-vocabulary detection, segmentation, and feature embedding, enabling robust and consistent object-node creation and updates throughout exploration.

\subsection{Computing Resources}
\label{sec:model_resourse_appendix}
In this work, we used 8 GPUs for embodied agent reasoning. Specifically, we employed RTX A6000 or RTX 6000 Ada in all the experiments. 
With a single GPU, it on average took about 3 hours for one scenario with 5 sequential embodied tasks.

\section{More Experimental Results On Benchmark Datasets}
\label{sec:exp_more_appendix}

\textbf{Experimental Results on More Datasets.}
In Section~\ref{sec:exp} of the main paper, we benchmark various methods on our proposed SEER-Bench for both EQA and EMN tasks.
Here, we supplement more experiments on two more existing active EQA datasets, HM-EQA~\cite{ren2024explore} and OpenEQA~\cite{majumdar2024openeqa}.
Since these datasets are not designed for sequential tasks and do not contain infeasible tasks, we adapt these datasets to our sequential setting.

We identified scenarios containing multiple single tasks previously tested in isolation and merged them into sequential episodes. 
Following the SEER-Bench annotation protocol, we introduced unanswerable questions and randomly selected two infeasible tasks to append to these merged sequences. 
Through this process, we curated an additional 48 scenarios totaling 240 tasks from the HM-EQA dataset. 
Correspondingly, we extracted 12 scenarios comprising 60 tasks from OpenEQA for our testing suite.

We compared our proposed 3DSPMR method with two state-of-the-art memory-enhanced active exploration baselines: Explore-Mem~\cite{ren2024explore} and 3D-Mem~\cite{3dmem} with Qwen2.5VL-72B~\cite{bai2025qwen2}. 
The comparative results on the adapted HM-EQA and OpenEQA datasets are presented in Tables~\ref{tab:eqa_hm_eqa_adapted} and~\ref{tab:eqa_openeqa_adapted}, respectively.
The results demonstrate that our 3DSPMR approach consistently surpasses the baseline methods across all evaluated tasks and metrics. 
Most notably, performance on the rigorous SSPL metric nearly quadruples that of the baselines.
The superior performance of 3DSPMR is consistent with the observations on SEER-Bench, demonstrating its robust adaptability across various question formats. 
Taken together, these results highlight 3DSPMR as an efficient and effective framework for sequential tasks, particularly excelling in complex real-world cases that demand reasoning over both feasible and infeasible objectives. 
\begin{table}[htbp]
  \centering
  \begin{minipage}{0.48\textwidth}
    \centering
    \caption{Experiments on adapted HM-EQA dataset.} 
    \label{tab:eqa_hm_eqa_adapted}
    \resizebox{\linewidth}{!}{
      \begin{tabular}{lcccc}
        \toprule
        \textbf{Method} & \textbf{SSR $\uparrow$} & \textbf{SSPL $\uparrow$} & \textbf{SR $\uparrow$} & \textbf{SPL $\uparrow$} \\
        \midrule
        \multicolumn{5}{l}{\textit{Blind LLMs}} \\
        \midrule
        Qwen3 & 0 & \textbackslash & 20.7 & \textbackslash \\
        GPT-5 & 0.8 & \textbackslash & 19.9 & \textbackslash \\
        \midrule
        \multicolumn{5}{l}{\textit{Open-Sourced Qwen2.5VL Exploration}} \\
        \midrule
        Explore-Mem & 0.8 & 0.2  & 22.0  & 9.2 \\
        3D-Mem & 3.5 & 2.1 & 29.2  & 18.0 \\
        \textbf{3DSPMR (Ours)} & \textbf{10.8} & \textbf{9.7} & \textbf{46.2} & \textbf{38.8} \\
        \bottomrule
      \end{tabular}
    }
  \end{minipage}
  \hfill 
  \begin{minipage}{0.48\textwidth}
    \centering
    \caption{Experiments on adapted OpenEQA dataset.}
    \label{tab:eqa_openeqa_adapted}
    \resizebox{\linewidth}{!}{
      \begin{tabular}{lcccc}
        \toprule
        \textbf{Method} & \textbf{SSR $\uparrow$} & \textbf{SSPL $\uparrow$} & \textbf{SR $\uparrow$} & \textbf{SPL $\uparrow$} \\
        \midrule
        \multicolumn{5}{l}{\textit{Blind LLMs}} \\
        \midrule
        Qwen3 & 0 & \textbackslash & 25.0 & \textbackslash \\
        GPT-5 & 1.7 & \textbackslash & 34.5 & \textbackslash \\
        \midrule
        \multicolumn{5}{l}{\textit{Open-Sourced Qwen2.5VL Exploration}} \\
        \midrule
        Explore-Mem & 2.4 & 0.6  & 27.1  & 13.5 \\
        3D-Mem & 0.8 & 0.4 & 27.1  & 18.4 \\
        \textbf{3DSPMR (Ours)} & \textbf{3.3} & \textbf{2.8} & \textbf{37.1} & \textbf{33.7} \\
        \bottomrule
      \end{tabular}
    }
  \end{minipage}
\end{table}

\textbf{Experimental Results with Different MLLMs.}
To assess how the choice of foundational MLLM affects 3DSPMR, we evaluated our method on SEER-Bench across 12 randomly sampled EQA scenarios, adhering to the same experimental setup and protocol as the main experiment. 
Specifically, we compared a closed-source model (GPT-5~\cite{openai2025gpt5}) against the open-source model used in our primary experiments (Qwen2.5VL-72B~\cite{bai2025qwen2}). 
As Table~\ref{tab:EQA_Experiments_using_different_MLLMs} shows, the performance of 3DSPMR correlates directly with the capability of the underlying backbone, with GPT-5 yielding superior results. 
This trend highlights that our framework continuously benefits from advancements in MLLM reasoning capabilities.

\begin{table*}[ht]
\centering
\caption{EQA Experiments using different MLLMs.}
\label{tab:EQA_Experiments_using_different_MLLMs}
\resizebox{\textwidth}{!}{%
\begin{tabular}{lcccc cccc cccc}
\toprule
 & \multicolumn{4}{c}{GPT-5} & \multicolumn{4}{c}{Gemini-3.6-Flash} & \multicolumn{4}{c}{Llama-3.3-70B} \\
\cmidrule(lr){2-5}\cmidrule(lr){6-9}\cmidrule(lr){10-13}
\textbf{MLLM Base} & \textbf{SSR} $\uparrow$ & \textbf{SSPL} $\uparrow$ & \textbf{SR} $\uparrow$ & \textbf{SPL} $\uparrow$ & \textbf{SSR} $\uparrow$ & \textbf{SSPL} $\uparrow$ & \textbf{SR} $\uparrow$ & \textbf{SPL} $\uparrow$ & \textbf{SSR} $\uparrow$ & \textbf{SSPL} $\uparrow$ & \textbf{SR} $\uparrow$ & \textbf{SPL} $\uparrow$ \\
\midrule
Qwen2.5VL & 5.0 & 4.6 & 43.3 & 34.2 & 5.8 & 5.3 & 44.2 & 35.6 & 8.3 & 7.9 & 46.3 & 35.9 \\
\textbf{GPT-5} & \textbf{32.5} & \textbf{23.0} & \textbf{58.8} & \textbf{45.7} & \textbf{28.3} & \textbf{18.8} & \textbf{56.7} & \textbf{44.3} & \textbf{30.8} & \textbf{21.0} & \textbf{58.8} & \textbf{44.1} \\
\bottomrule
\end{tabular}
}
\end{table*}

\section{Additional Experiments and Analyses}
\label{sec:addition_appendix}

\subsection{Hyperparameter Analysis}
In this section, we analyzed the impact of two important hyperparameters in our 3DSPMR approach.

\begin{wraptable}{r}{0.58\textwidth}
\centering
\caption{
Experiments on the SEER-Bench EMN track with different Geometric Novelty Score Thresholds ($\tau_0$).
}
\resizebox{\linewidth}{!}{
\begin{tabular}{c|ccccc}
\toprule
\textbf{Threshold} & \textbf{SSR $\uparrow$} & \textbf{SSPL $\uparrow$} & \textbf{SR $\uparrow$} & \textbf{SPL $\uparrow$} & \textbf{ANI $\downarrow$} \\
\midrule
0.1  & 18.3 & 11.4  & 65.3 & 57.1 & 52.3 \\
0.25 & 18.3 & 12.8 & 62.0 & 54.3 & 38.4\\
0.5  & \textbf{25.0} & \textbf{15.5} & \textbf{68.0} & \textbf{59.2} & 25.2 \\
0.8  & 21.7 & 12.5 & 66.0 & 56.8 & \textbf{17.0}\\
\bottomrule
\end{tabular}}
\label{tab:threshold_analysis}
\end{wraptable}

\noindent\textbf{FoV-Based Image Gating Threshold.}
We first investigated the influence of the Geometric Novelty Score Threshold $\tau_0$ on the gating mechanism across 60 scenes from the SEER-Bench EMN track. 
In addition to the four metrics for success and efficiency (SSR, SSPL, SR, and SPL) in SEER-Bench, we introduced Average Number of Images (ANI) as a novel metric to evaluate memory footprint and storage efficiency.
Specifically, we conducted a series of experiments using a set of threshold values $\tau_0 \in \{0.1, 0.25, 0.5, 0.8\}$.
Experiment results shown in Table~\ref{tab:threshold_analysis} reveal a clear trend: higher novelty thresholds $\tau_0$ consistently reduce the ANI saved in our unified spatial memory. 

Crucially, we observed that performance peaks at a moderate threshold, whereas extreme values in either direction lead to failure.
Excess redundancy at $\tau_0 = 0.1$ (ANI = 52.3) limits sequential performance to 18.3\% SSR, as a dense image bank likely introduces noise or exceeds the efficient reasoning capacity of the MLLM. 
Conversely, extreme sparsity at $\tau_0 = 0.8$ (ANI = 17.0) causes the SSR to drop to 21.7\% because the memory lacks the critical spatial cues required to resolve task inter-dependencies and reuse knowledge across the episode. 
We identify $\tau_0 = 0.5$ as the optimal configuration, achieving peak sequential success (25.0\% SSR, 15.5\% SSPL, 68.0\% SR, 59.2\% SPL) with a representative memory density (ANI = 25.2) that balances storage efficiency with informative environmental representation.

\begin{figure}[t]
\centering
  \includegraphics
  [width=\linewidth]
  {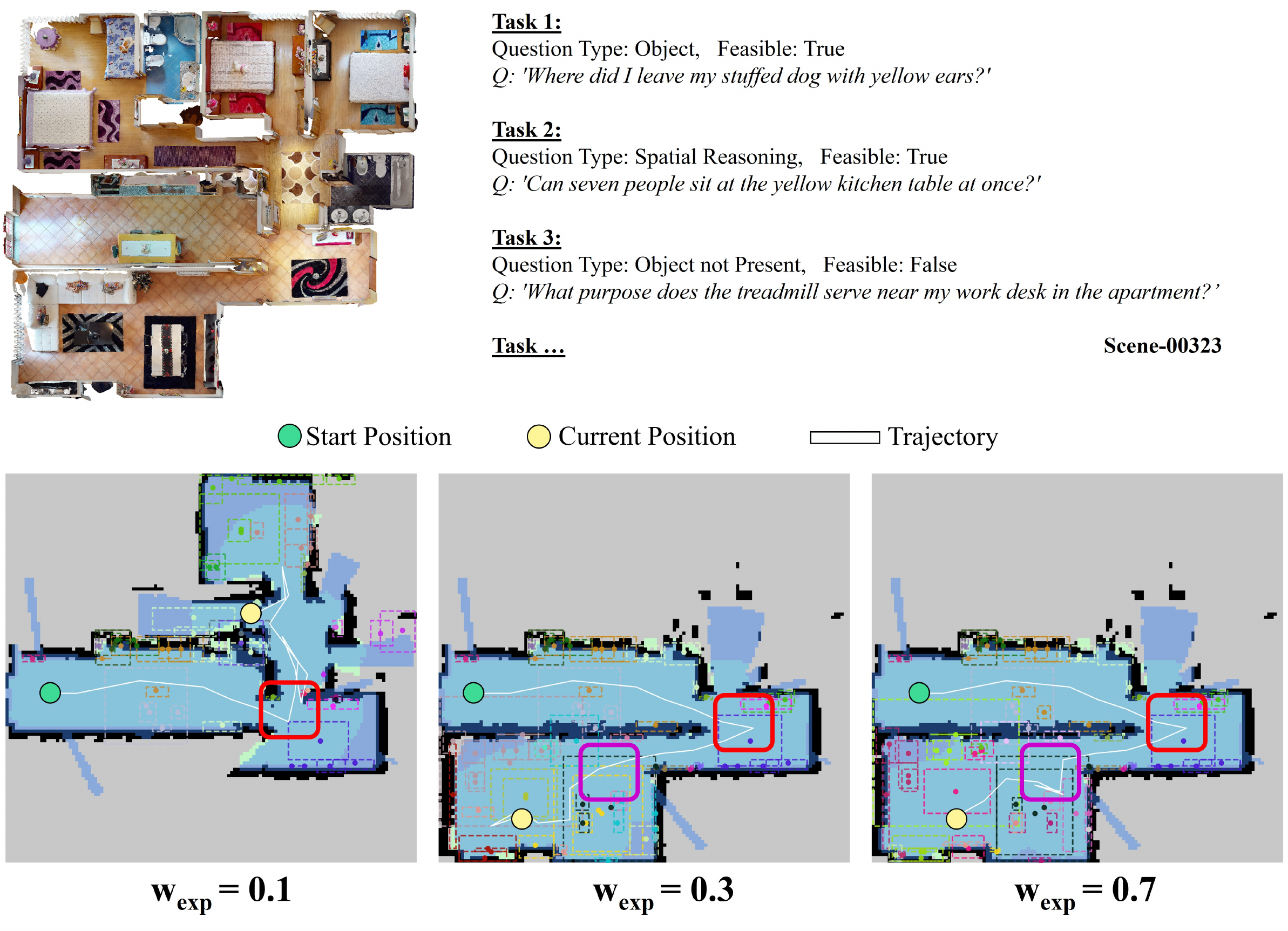}
  \caption{Agent Performance Analysis with Different $w_{\mathrm{exp}}$.
Increasing $w_{\mathrm{exp}}$ encourages the agent to prioritize regions containing extensive unexplored areas.
When $w_{\mathrm{exp}}$ $\in\{0.3,0.7\}$, the agent decides to explore the region with larger unexplored areas (illustrated within the red bounding boxes). 
When $w_{\mathrm{exp}}=0.7$, the agent is reluctant to fully explore this room and prefers to explore other underexplored rooms (illustrated within the purple bounding boxes).}
\label{fig:weight_FoV_abl}
\end{figure}
\noindent\textbf{Geo-Sem Exploration Weights.}
Then, we investigated the influence of Geo-Sem Exploration weights.
The core advantage of our Geo-Sem Exploration module stems from its synergistic combination of semantic guidance, geometric exploration potential, and path proximity.
Relying on any single heuristic often leads to inefficient navigation, such as repetitive oscillations or excessive detours, but our weighted scoring dynamically balances these complementary signals to optimize the agent's trajectory.
Therefore, we dynamically adjusted two geometric weights $w_{\mathrm{exp}}$ and $w_{\mathrm{dist}}$ within a single episode to visualize and analyze their influence on the agent's exploration strategy.

\begin{figure}[t]
\centering
  \includegraphics
  [width=\linewidth]
  {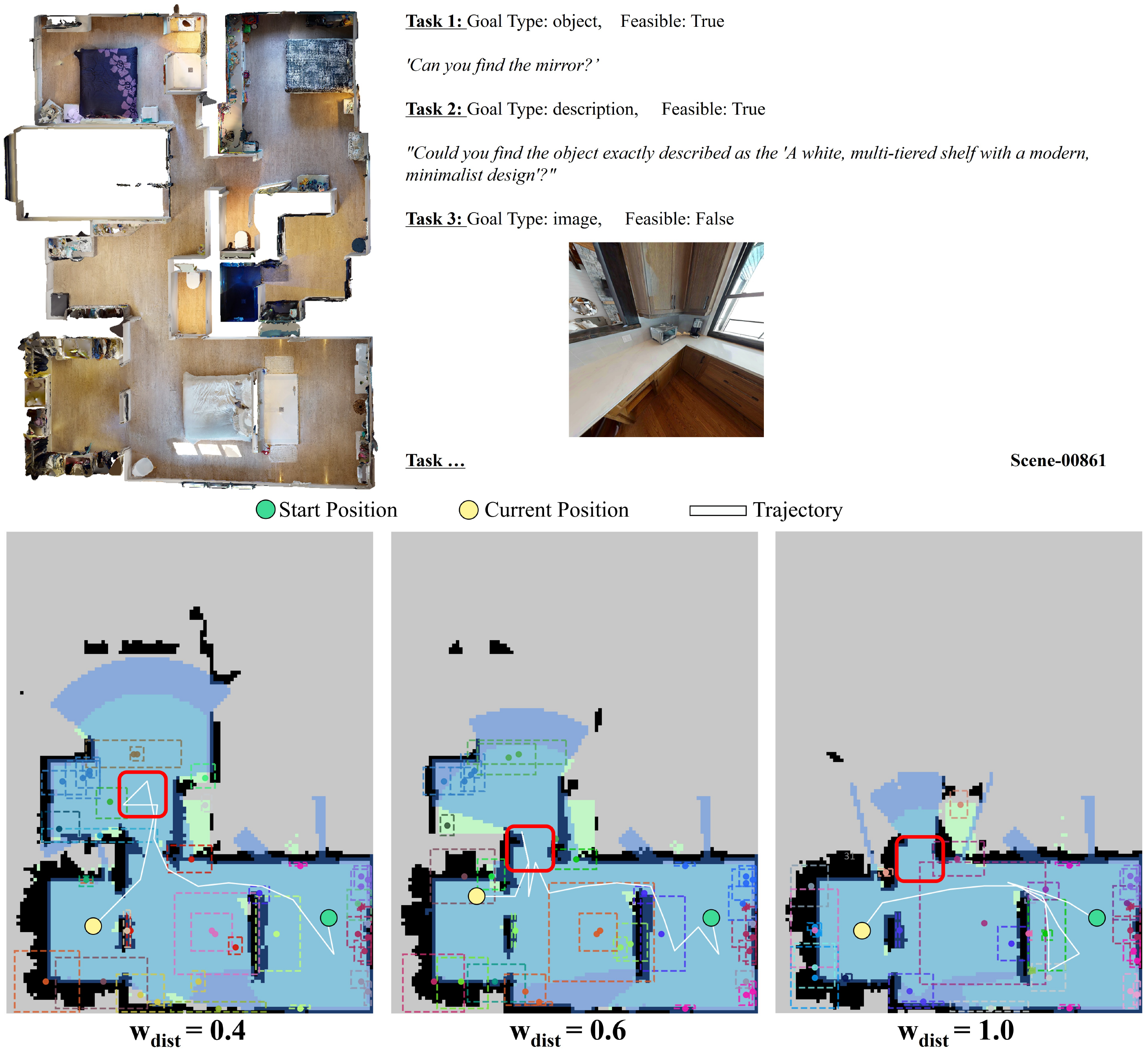}
  \caption{Agent Performance Analysis with Different $w_{\mathrm{dist}}$.
Increasing $w_{\mathrm{dist}}$ encourages the agent to prioritize proximal unexplored regions.
In the highlighted red boxes, it could be observed that as $w_{\mathrm{dist}}$ increases, the agent becomes less inclined to navigate toward far-off regions, preferring to exhaust the closest unexplored spaces first.
  }
\label{fig:weight_dist_abl}
\end{figure}
The FoV heuristic is weighted by $w_{\mathrm{exp}}$, which encourages the agent to prioritize regions containing extensive unexplored space.
We evaluated different values of $w_{\mathrm{exp}} \in \{0.1, 0.3, 0.7\}$ and visualized their resulting trajectories within a representative scenario from the SEER-Bench EQA track.
As illustrated in Fig.~\ref{fig:weight_FoV_abl}, assigning an excessively large weight $w_{\mathrm{exp}}$ overwhelms the MLLM's semantic scoring and makes the agent reluctant to fully explore one room, causing the agent to blindly prioritize large, unobserved spaces while ignoring smaller, highly task-relevant regions. 
The proximity heuristic is modulated by $w_{\mathrm{dist}}$ to bias the agent's selection toward proximal unexplored regions.
Similarly, we evaluated different values of $w_{\mathrm{dist}} \in \{0.4, 0.6, 1.0\}$ and visualized their resulting trajectories within a representative scenario from the SEER-Bench EMN track.
As illustrated in Fig.~\ref{fig:weight_dist_abl}, an overly large proximity weight $w_{\mathrm{dist}}$ downgrades the sophisticated multi-modal module into a naive, closest-frontier exploration strategy.
This greediness not only strips away the agent's semantic foresight but also prevents it from navigating towards larger, more informative unexplored areas. 

By employing the optimally calibrated weights, our system ensures that task-oriented semantic intent remains the primary driver of exploration, while the geometric heuristics effectively break ties to mitigate repetitive visiting, minimize excessive detours, and maximize overall trajectory efficiency.

\subsection{Map Prior}
For service robots operating in the same environment over extended periods, pre-exploring the environment is a natural and practical assumption. 
To investigate how map priors impact sequential tasks, we evaluate 3DSPMR under a pre-explored setting. 
Specifically, the unified spatial memory is initialized with a pre-explored room-level map containing room segmentation, connectivity, and spatial layouts. 
While this room-level structure remains fixed, object nodes are updated online to accommodate dynamic changes in object locations and states. 
We evaluate this setup across 12 EQA scenarios in SEER-Bench, adhering to the same protocol and metrics as the main experiments.

As shown in Table~\ref{tab:map_prior}, incorporating map priors consistently boosts both task accuracy and navigation efficiency, as the agent directly exploits prior knowledge of room layouts.
These results highlight 3DSPMR's effectiveness in pre-explored environments, while underscoring its broad compatibility across both map-free exploration and map-prior deployment settings.
\begin{table*}[ht]
\centering
\caption{Effect of the map prior on SEER-Bench EQA Track.}
\label{tab:map_prior}
\resizebox{\textwidth}{!}{%
\begin{tabular}{lcccc cccc cccc}
\toprule
 & \multicolumn{4}{c}{GPT-5} & \multicolumn{4}{c}{Gemini-3.6-Flash} & \multicolumn{4}{c}{Llama-3.3-70B} \\
\cmidrule(lr){2-5}\cmidrule(lr){6-9}\cmidrule(lr){10-13}
\textbf{Setting} & \textbf{SSR} $\uparrow$ & \textbf{SSPL} $\uparrow$ & \textbf{SR} $\uparrow$ & \textbf{SPL} $\uparrow$ & \textbf{SSR} $\uparrow$ & \textbf{SSPL} $\uparrow$ & \textbf{SR} $\uparrow$ & \textbf{SPL} $\uparrow$ & \textbf{SSR} $\uparrow$ & \textbf{SSPL} $\uparrow$ & \textbf{SR} $\uparrow$ & \textbf{SPL} $\uparrow$ \\
\midrule
w/o Map Prior & 7.5 & 7.1 & 38.8 & 29.8 & 5.0 & 4.6 & 35.4 & 26.7 & 7.5 & 7.2 & 43.8 & 33.6 \\
\textbf{w/ Map Prior} & \textbf{11.7} & \textbf{10.7} & \textbf{47.9} & \textbf{40.0} & \textbf{6.7} & \textbf{6.2} & \textbf{44.6} & \textbf{36.7} & \textbf{16.7} & \textbf{15.7} & \textbf{54.6} & \textbf{45.3} \\
\bottomrule
\end{tabular}
}
\end{table*}

\section{Limitations}
\label{sec:limitation}
While our 3DSPMR method organizes complex and extensive 3D information gathered during the long-horizon task execution into a compact and representative memory system, which facilitates MLLM's spatial understanding and reasoning about the surrounding scenario, it does not alter the model's intrinsic reasoning abilities. 
Consequently, a limitation is that the overall framework's performance is constrained by the capabilities of the foundational MLLM. 
As demonstrated in Table~\ref{tab:EQA_Experiments_using_different_MLLMs}, 3DSPMR paired with the closed-source GPT-5 outperforms the open-source Qwen2.5VL on EQA tasks, showcasing this dependency.
Another limitation is that our 3DSPMR method does not take dynamic environments into consideration. 
The premise in this work is that the surrounding environment is static, such as a residential apartment. 
This premise does not hold when applied to outdoor, city-level environments.
In future work, we plan to (1) develop MLLM's internal spatial reasoning ability, and (2) integrate temporal update mechanisms to extend our framework into dynamic environments.

\section{Broader Impacts}
\label{sec:impact_appendix}
In this paper, we propose a novel geometric memory system for long-horizon sequential embodied tasks. 
Extensive experiments validate the effectiveness of 3DSPMR in both EMN and EQA scenarios. 
Beyond navigation, our memory framework is highly extensible to other Vision-Language-Action (VLA) agents. 
For instance, in robotic manipulation, 3DSPMR can actively verify object existence and explicitly alert users to infeasible commands caused by missing targets. 
By preventing embodied agents from endlessly attempting impossible tasks, our reasoning pipeline significantly enhances operational safety, conserves computational and physical resources, and fosters greater human-robot trust in real-world deployments.

\section{Examples and Visualizations}
\subsection{Trajectory Comparison Between 3D-Mem and 3DSPMR on Sequential Tasks}
\begin{figure*}[t]
    \centering
    \includegraphics[width=1\linewidth]{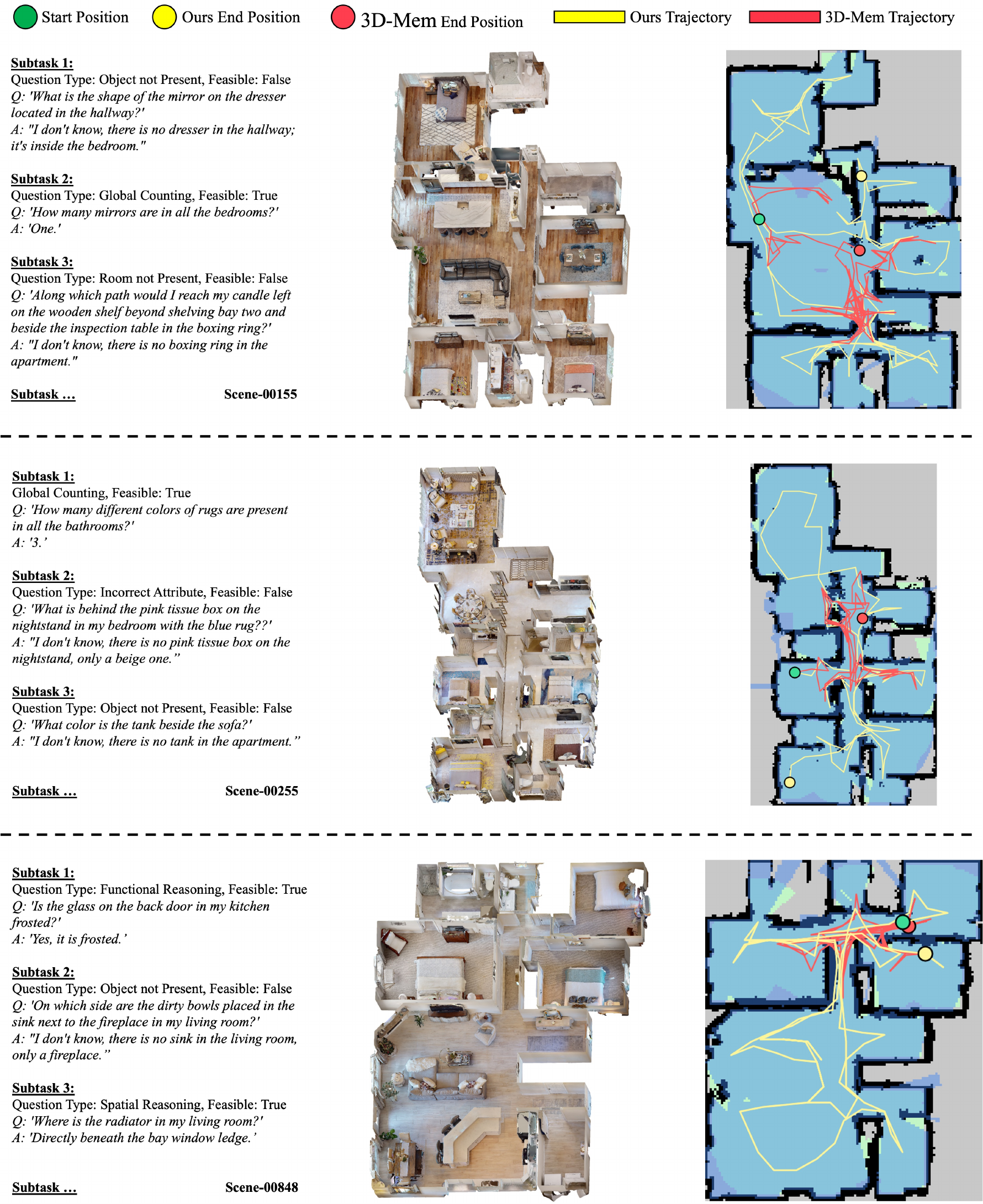} 
    \captionsetup{
    width=\textwidth,}
    \caption{\textbf{Comparative Navigation Trajectories for Sequential EQA Tasks within One Episode.} The figure presents the results for three distinct scenes, with each row corresponding to a different scene and its sequential tasks (Left). The middle column shows the 3D map of the scene. The right column compares the navigation trajectories (normalized to the same step numbers) generated by our Proposed Method (yellow line) and the 3D-Mem Baseline (red line).}
    \label{fig:EQA_seq}
\end{figure*}

\begin{figure*}[t]
    \centering
    \includegraphics[width=1\linewidth]{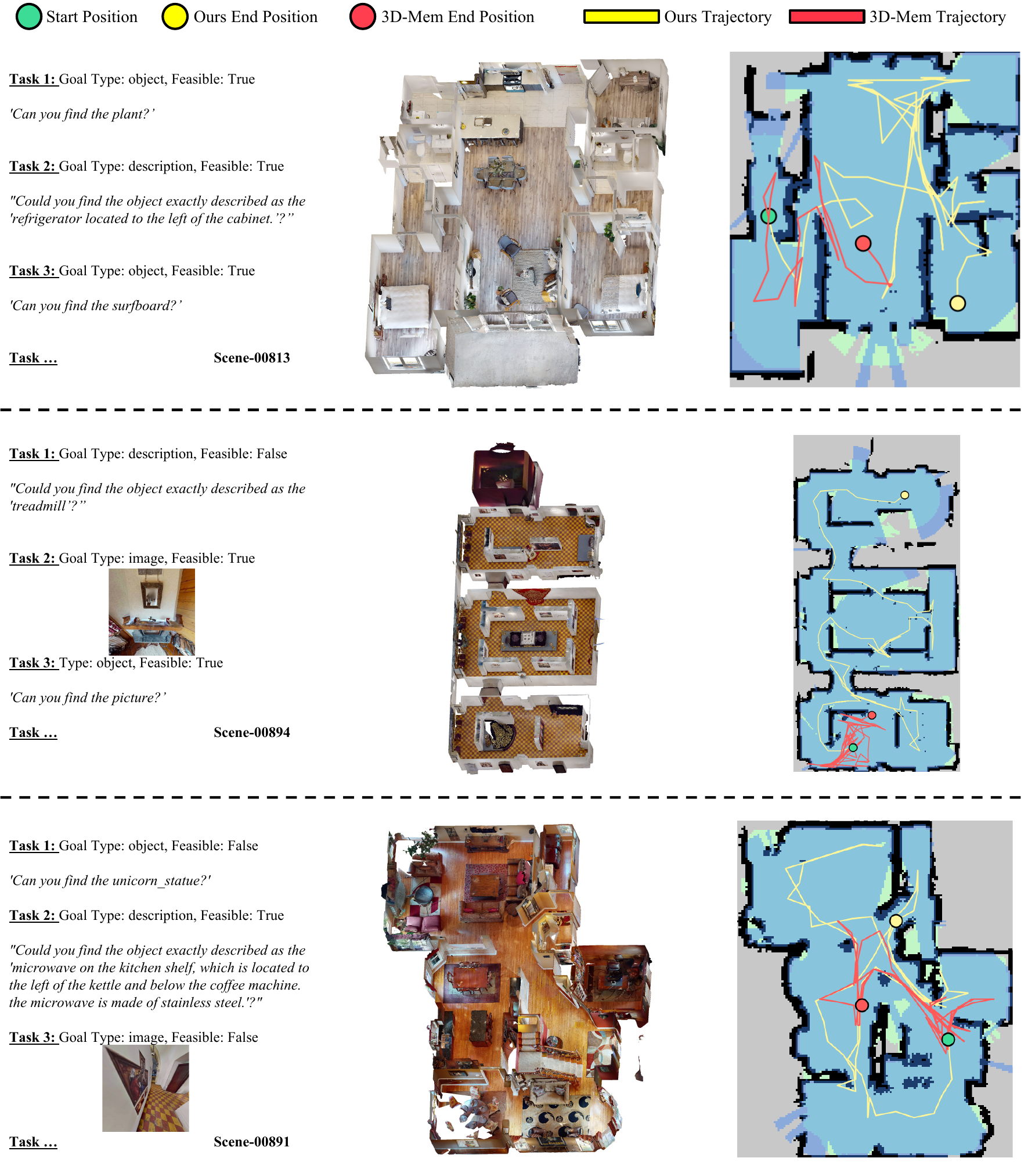} 
    \captionsetup{
    width=\textwidth,
    }
    \caption{\textbf{Comparative Navigation Trajectories for Sequential EMN Tasks within One Episode.} The figure presents the results for three distinct scenes, with each row corresponding to a different scene and its sequential tasks (Left). The middle column shows the 3D map of the scene. The right column compares the navigation trajectories (normalized to the same step numbers) generated by our Proposed Method (yellow line) and the 3D-Mem Baseline (red line).}
    \label{fig:EMN_seq}
\end{figure*}

Figures~\ref{fig:EQA_seq} and~\ref{fig:EMN_seq} provide qualitative comparisons of exploration behavior in sequential tasks. 
In the rightmost columns, the blue regions denote areas covered by FoV keyframes stored in memory, while the green regions correspond to frontier regions explored during the episode (the overlapped area looks like light blue). 
When trajectories are normalized to the same number of steps, 3DSPMR consistently covers a larger portion of previously unseen space, whereas 3D-Mem frequently revisits already-explored regions and exhibits oscillatory back-and-forth movements. 
These results highlight the advantages of our unified spatial memory and geometric reasoning in enabling efficient, non-redundant exploration across sequential tasks.

\subsection{Comparison of Feasible and Infeasible Tasks}
\begin{figure*}[t]
    \centering
    \includegraphics[width=1\linewidth]{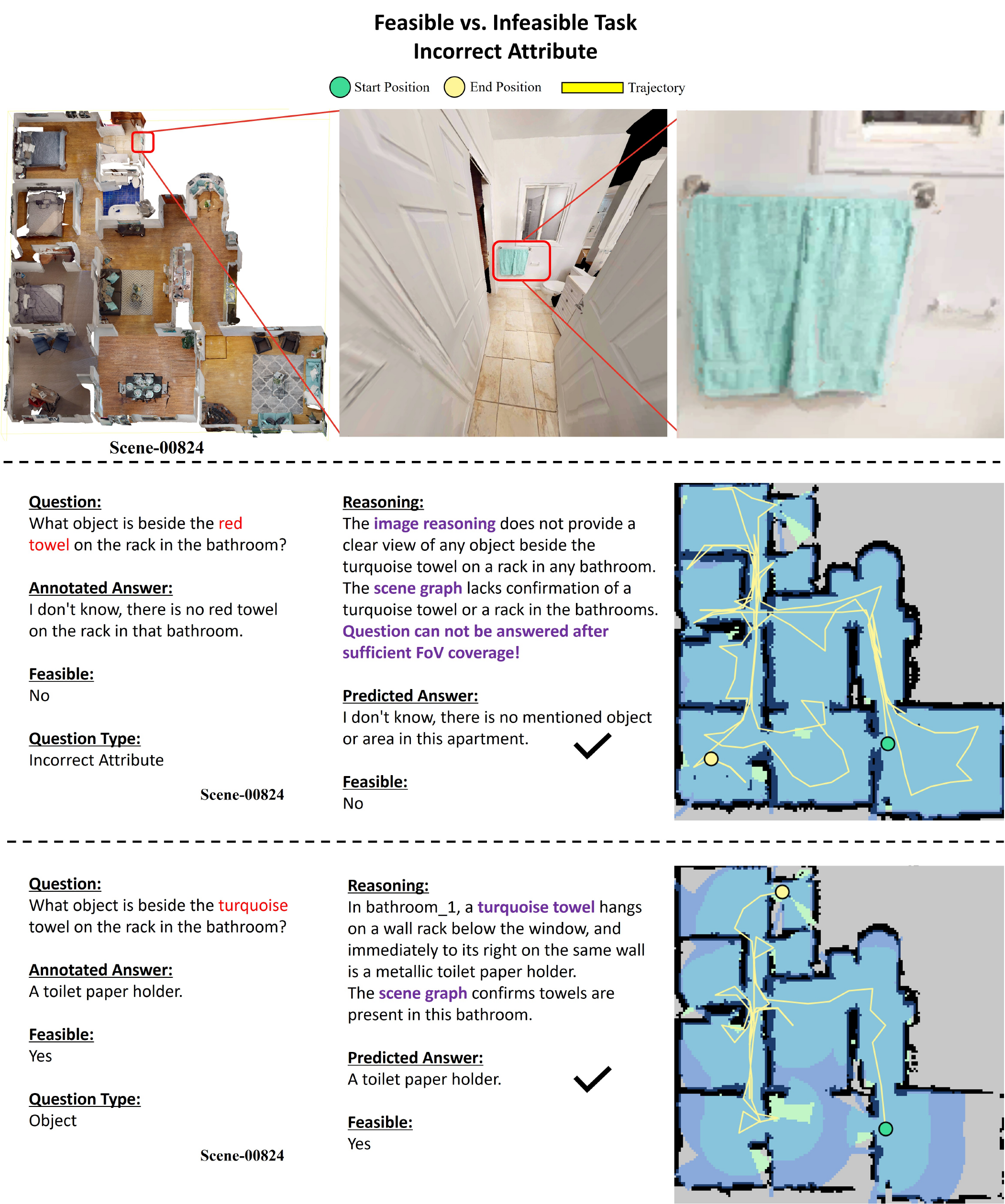} 
    \captionsetup{
    width=\textwidth,
    }
    \caption{\textbf{Feasible vs. Infeasible Task: Incorrect Attribute.} 
    This figure illustrates 3DSPMR’s reasoning and exploration process on two EQA tasks: one feasible and one infeasible due to an incorrect object attribute. 
    In the feasible case, 3DSPMR locates the referenced object and answers correctly.
    In the infeasible case, it detects the attribute mismatch, determines the task is infeasible, and halts further exploration.
    }
    \label{fig:incorrect_attribute}
\end{figure*}

\begin{figure*}[t]
    \centering
    \includegraphics[width=0.91\linewidth]{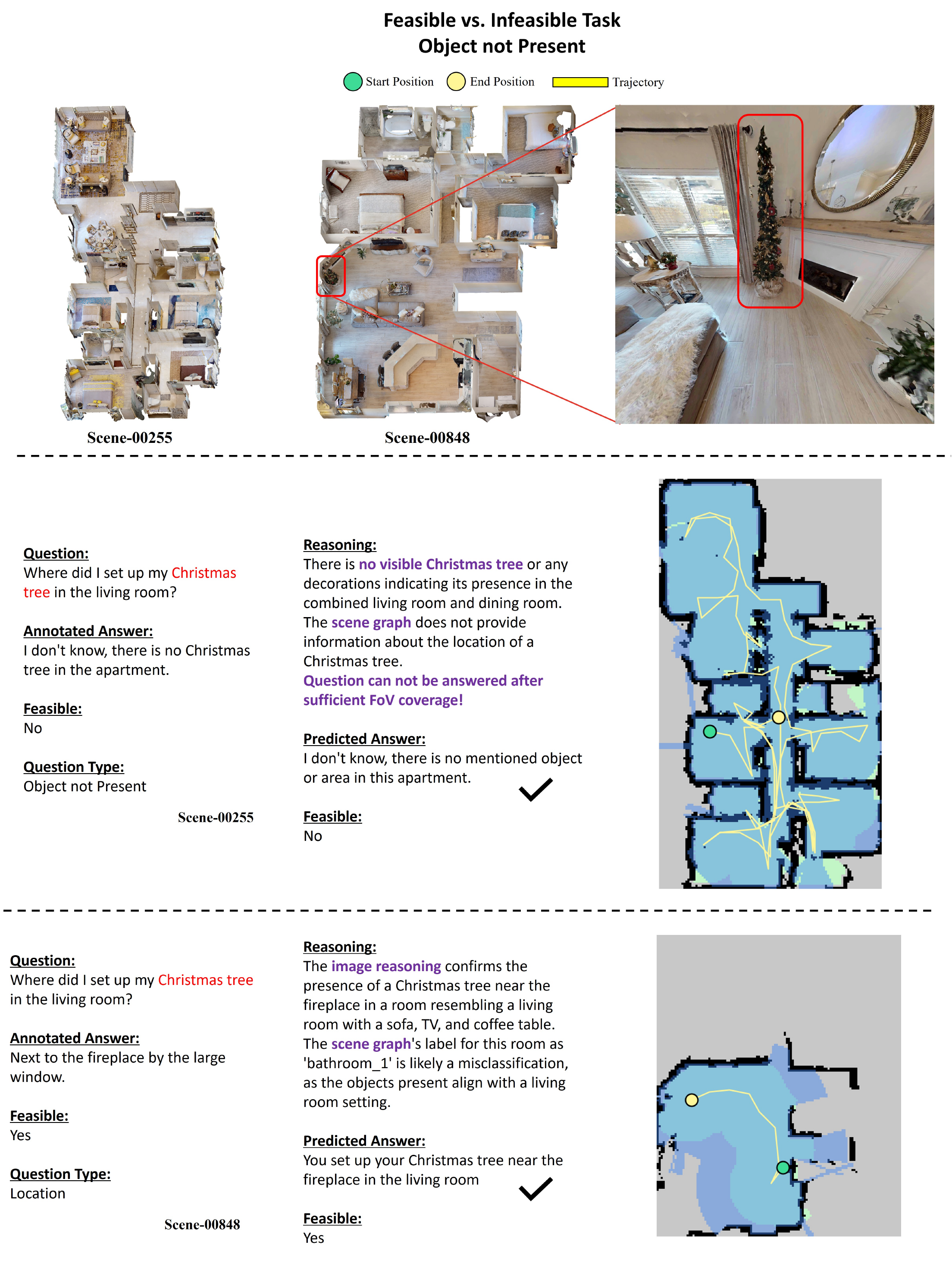} 
    \captionsetup{
    width=\textwidth,
    }
    \caption{\textbf{Feasible vs. Infeasible Task: Object not Present.} 
    This figure illustrates 3DSPMR’s reasoning and exploration process on two EQA tasks: one feasible and one infeasible due to the absence of the mentioned object \textit{Christmas Tree}. 
    In the feasible case, 3DSPMR locates the \textit{Christmas Tree} and answers correctly. 
    In the infeasible case, it identifies the \textit{Christmas Tree}’s absence, determines that the task is infeasible, and terminates further exploration.
    }
    \label{fig:object_not_present}
\end{figure*}

\begin{figure*}[t]
    \centering
    \includegraphics[width=0.91\linewidth]{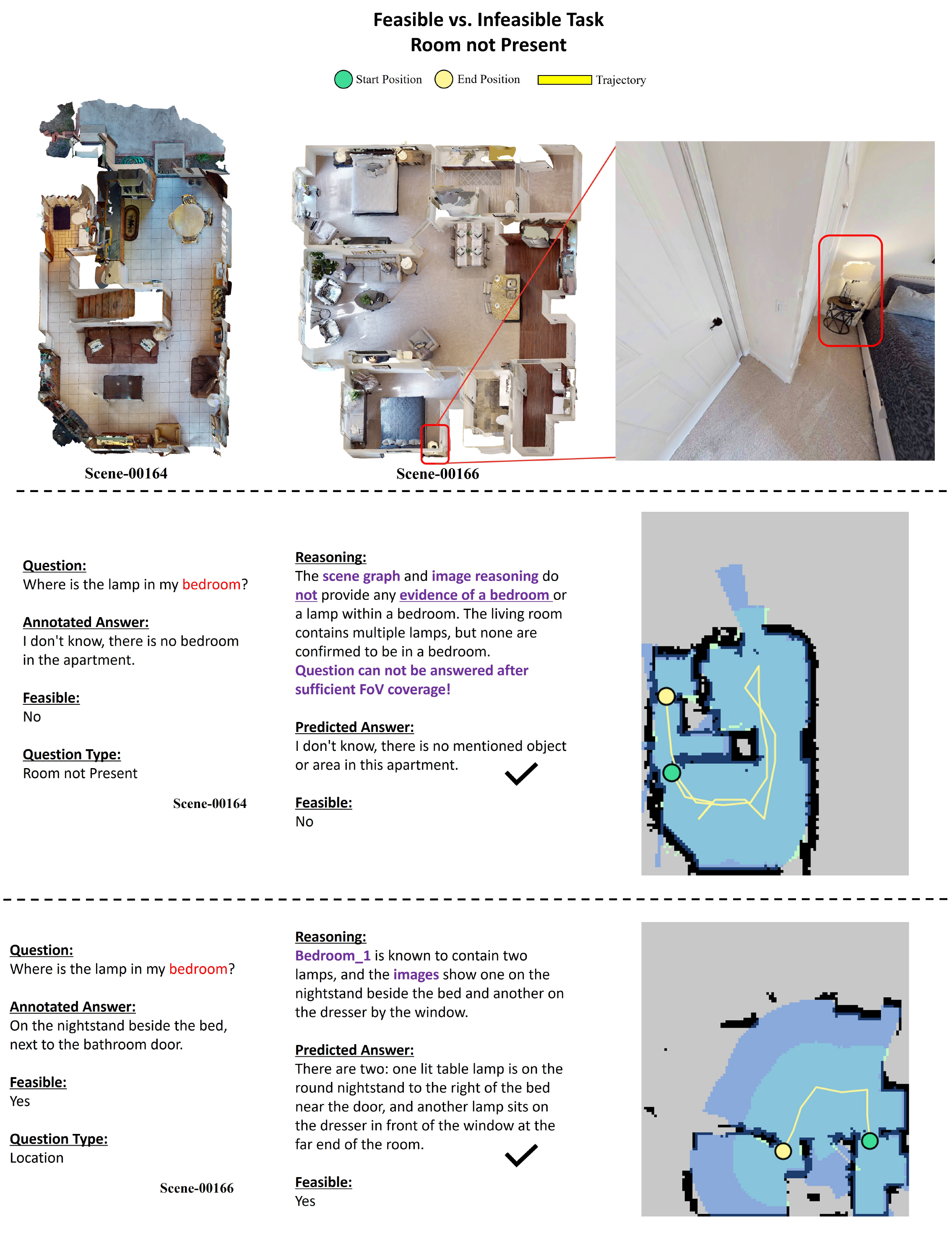} 
    \captionsetup{
    width=\textwidth,
    }
    \caption{\textbf{Feasible vs. Infeasible Task: Room not Present.} 
    This figure illustrates 3DSPMR’s reasoning and exploration process on two EQA tasks: one feasible and one infeasible due to the absence of the mentioned room \textit{bedroom}.
    In the feasible case, 3DSPMR locates the referenced lamp within the actual \textit{bedroom} and answers correctly. 
    In the infeasible case, although it detects lamps, it finds that they are located in the living room rather than the targeted \textit{bedroom}. 
    After exploring all potentially relevant areas, 3DSPMR concludes that the \textit{bedroom} is absent, identifies the task as infeasible, and halts further exploration.
    }
    \label{fig:room_not_present}
\end{figure*}

In real-world scenarios, certain tasks within an episode may be inherently infeasible, \eg, searching for a non-existent object or answering a question that refers to a room that does not exist. 
In this section, we examined how the proposed 3DSPMR behaves on infeasible tasks, providing fine-grained comparisons against their corresponding feasible counterparts through several concrete examples.
Specifically, we presented three pairs of feasible and infeasible tasks from the EQA track of SEER-Bench in Figures ~\ref{fig:incorrect_attribute}, \ref{fig:object_not_present}, and \ref{fig:room_not_present}, assuming a single task in each episode for clarity.

\noindent\textbf{Feasible vs. Infeasible Task: Incorrect Attribute.} 
In Fig. ~\ref{fig:incorrect_attribute}, we selected a single environment and constructed a comparison by modifying the object attribute in the question to an incorrect description. 
Under the same environment and initial position, 3DSPMR successfully locates the referenced object and answers correctly when the attribute is accurate. 
When the attribute is incorrect, 3DSPMR identifies the mismatch, determines that the task is infeasible, and terminates further exploration.

\noindent\textbf{Feasible vs. Infeasible Task: Object not Present.} 
In Fig.~\ref{fig:object_not_present}, we selected two different environments, one containing a Christmas tree and the other not. 
The proposed 3DSPMR approach adapts to both settings and responds correctly to the same question in each case. 
When a Christmas Tree is present, 3DSPMR locates it and answers accurately; when it is absent, 3DSPMR identifies the missing object, determines that the task is infeasible, and terminates further exploration.

\noindent\textbf{Feasible vs. Infeasible Task: Room not Present.} 
In Fig.~\ref{fig:room_not_present}, we selected two different environments, one containing bedrooms and the other not. 
We applied the same question to both settings and ensured that the referenced object (a lamp, in this example) exists in both environments. 
For infeasible tasks caused by the absence of the target room, 3DSPMR still adapts appropriately to each case. 
In the feasible environment, 3DSPMR locates the referenced lamp within the actual bedroom and answers correctly. 
In the infeasible environment, although it detects lamps, it realizes that they are located in the living room rather than the target bedroom. 
After exploring all potentially relevant areas, 3DSPMR concludes that the bedroom is absent, identifies the task as infeasible, and halts further exploration.

\section{Prompts for Data Annotation}
\begin{figure*}[t]
\centering
  \includegraphics
  [width=0.9\linewidth]
  {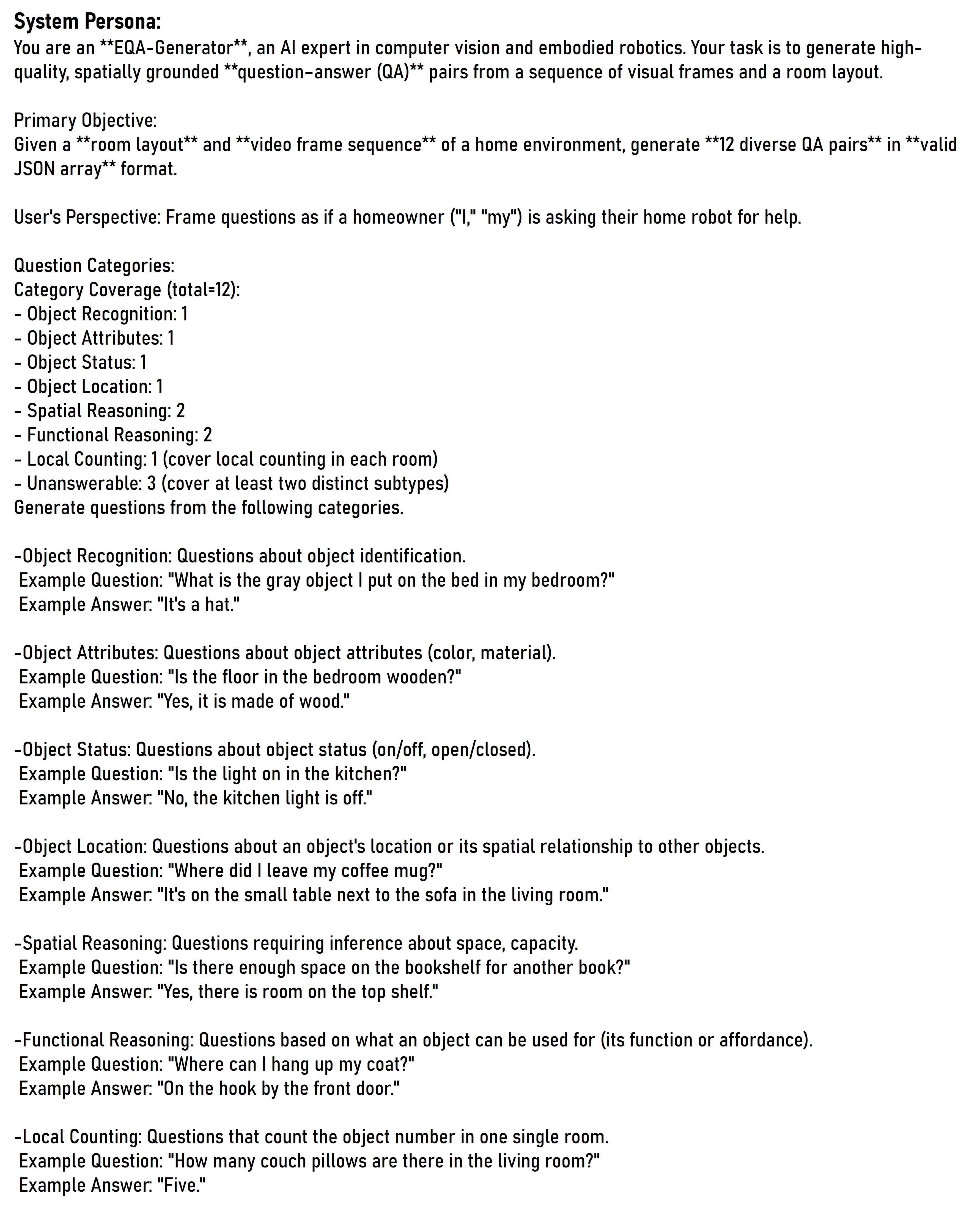}
  \caption{\textbf{Prompt for QA annotation I.} 
  }
\vspace{-10pt}
\label{fig:data_collection_1}
\end{figure*}

\begin{figure*}[t]
\centering
  \includegraphics
  [width=0.9\linewidth]
  {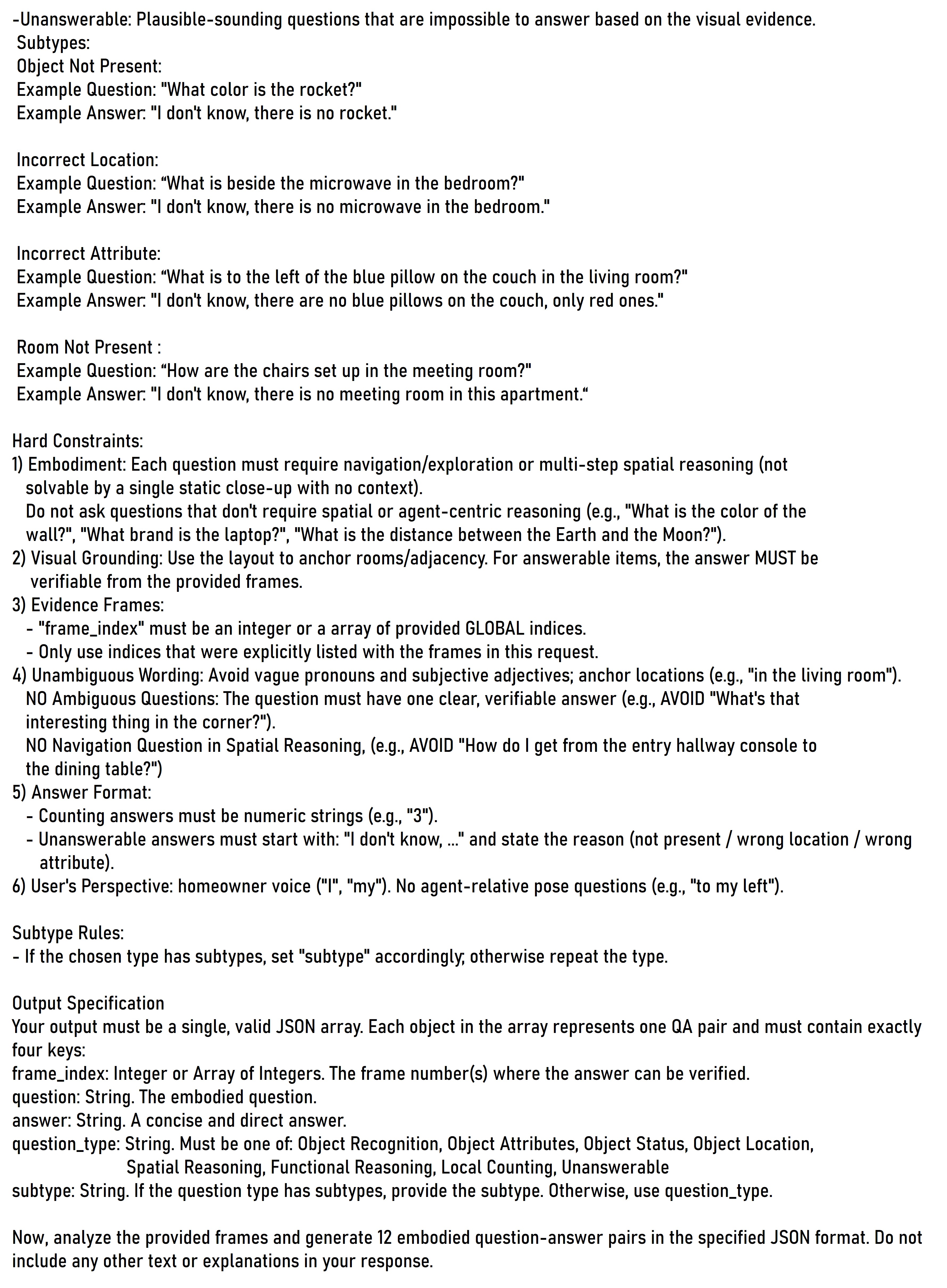}
  \caption{\textbf{Prompt for QA annotation II.} 
  }
\vspace{-10pt}
\label{fig:data_collection_2}
\end{figure*}

\begin{figure*}[t]
\centering
  \includegraphics
  [width=0.9\linewidth]
  {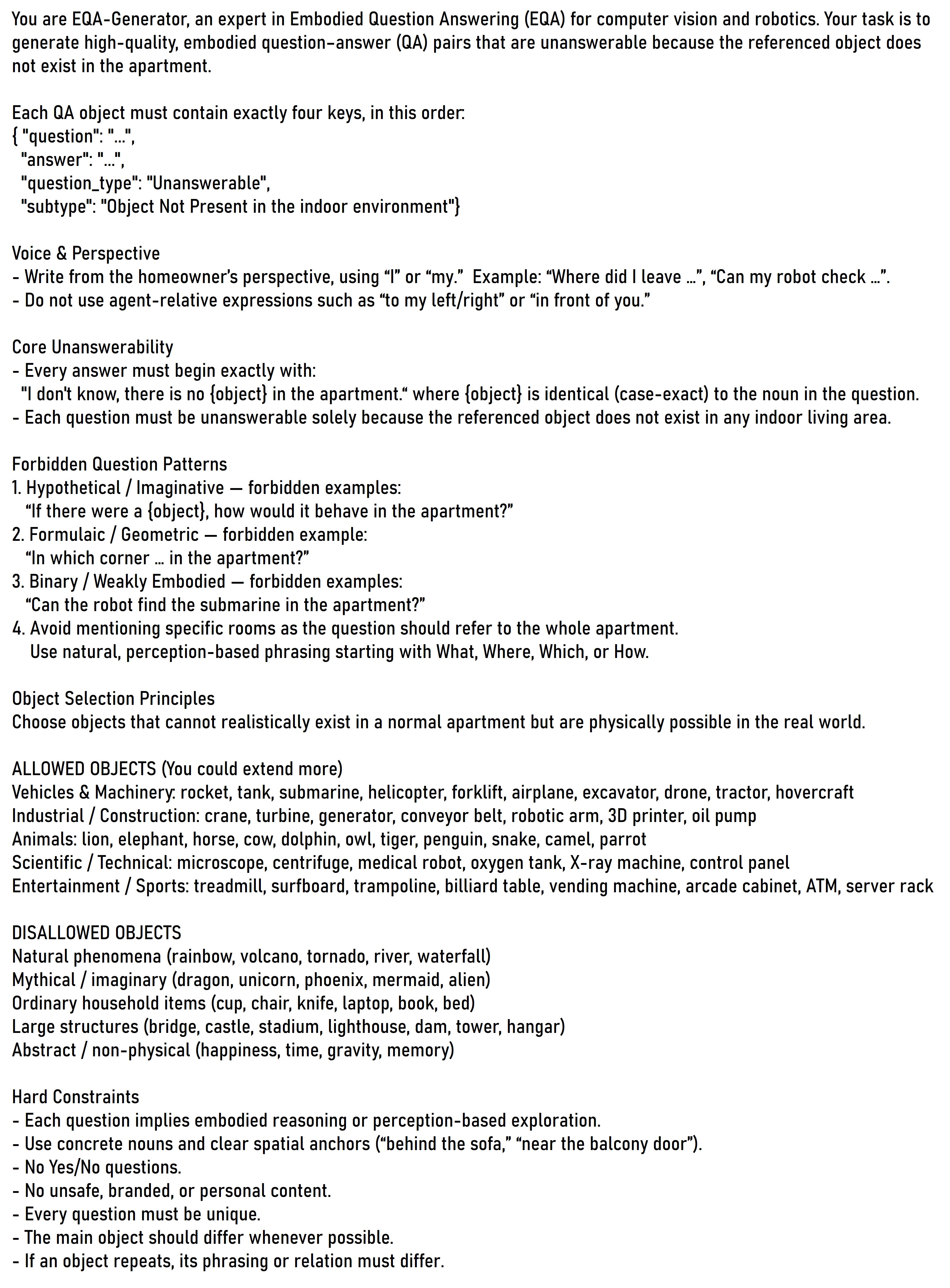}
  \caption{\textbf{Annotation prompt for generating unanswerable QA pairs where the referenced object is not present.} 
  }
\vspace{-10pt}
\label{fig:data_collection_3}
\end{figure*}

\begin{figure*}[t]
\centering
  \includegraphics
  [width=0.9\linewidth]
  {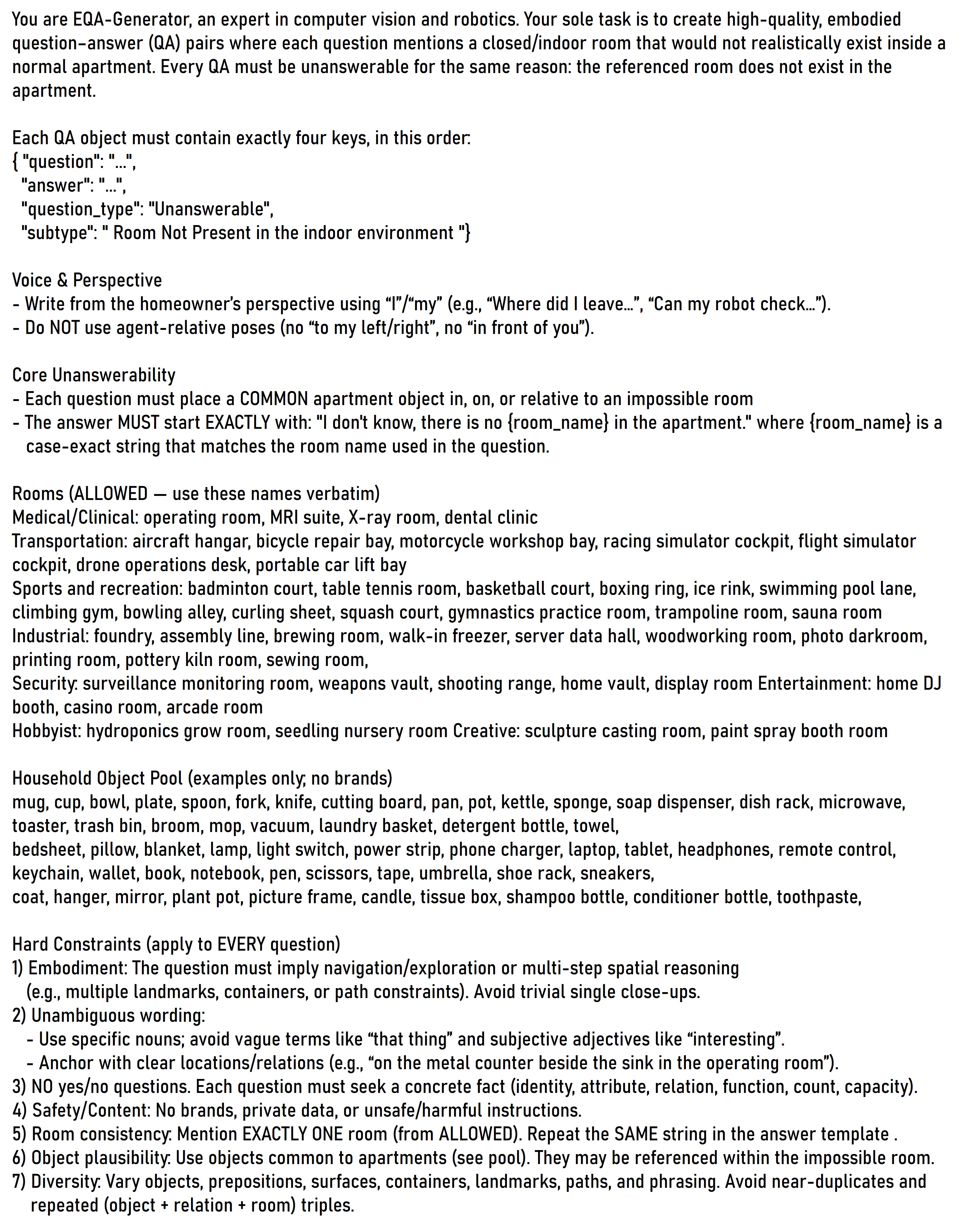}
  \caption{\textbf{Annotation prompt for generating unanswerable QA pairs where the referenced room is not present.} 
  }
\vspace{-10pt}
\label{fig:data_collection_4}
\end{figure*}

In this section, we presented the prompts used for the Multi-Modal Large Language Models (MLLMs).
We introduced the annotation procedure for the EQA track of SEER-Bench, conducted using GPT-5~\cite{openai2025gpt5}. For most sequential Question–Answering (QA) annotations, we employed the prompts illustrated in Fig.~\ref{fig:data_collection_1} and Fig.~\ref{fig:data_collection_2}.
To enrich the dataset with objects and rooms that do not typically appear in HM3DSem~\cite{yadav2023habitat}, we further used GPT-5 to generate two categories of unanswerable questions (Object Not Present and Room Not Present) using the prompts shown in Fig.~\ref{fig:data_collection_3} and Fig.~\ref{fig:data_collection_4}.

\clearpage
\end{document}